%% file: main.tex
\documentclass[acmsmall,screen,authorversion,nonacm]{acmart}

\AtBeginDocument{%
  \providecommand\BibTeX{{%
    \normalfont B\kern-0.5em{\scshape i\kern-0.25em b}\kern-0.8em\TeX}}}

\usepackage{graphicx}
\usepackage{amsmath}
\usepackage{url}
\usepackage{amsfonts}
\usepackage{multirow}
\usepackage{xcolor}
\usepackage{graphicx}
\usepackage{hyperref}
\mathchardef\mhyphen="2D

\begin{document}

\title{A Practical Survey on Faster and Lighter Transformers}

\author{Quentin~Fournier}
\email{quentin.fournier@polymtl.ca}
\affiliation{%
  \institution{Polytechnique Montréal}
  \streetaddress{2500 Chemin de Polytechnique}
  \city{Montréal}
  \state{Quebec}
  \country{Canada}
  \postcode{H3T 1J4}
}

\author{Gaétan~Marceau~Caron}
\email{gaetan.marceau.caron@mila.quebec}
\affiliation{%
  \institution{Mila - Quebec AI Institute}
  \streetaddress{6666 Rue Saint-Urbain}
  \city{Montréal}
  \state{Quebec}
  \country{Canada}
  \postcode{H2S 3H1}
}

\author{Daniel~Aloise}
\email{daniel.aloise@polymtl.ca}
\affiliation{%
  \institution{Polytechnique Montréal}
  \streetaddress{2500 Chemin de Polytechnique}
  \city{Montréal}
  \state{Quebec}
  \country{Canada}
  \postcode{H3T 1J4}
}

\renewcommand{\shortauthors}{Fournier et al.}

\begin{abstract}
  Recurrent neural networks are effective models to process sequences. However, they are unable to learn long-term dependencies because of their inherent sequential nature. As a solution, Vaswani et al. introduced the Transformer, a model solely based on the attention mechanism that is able to relate any two positions of the input sequence, hence modelling arbitrary long dependencies. The Transformer has improved the state-of-the-art across numerous sequence modelling tasks. However, its effectiveness comes at the expense of a quadratic computational and memory complexity with respect to the sequence length, hindering its adoption. Fortunately, the deep learning community has always been interested in improving the models' efficiency, leading to a plethora of solutions such as parameter sharing, pruning, mixed-precision, and knowledge distillation. Recently, researchers have directly addressed the Transformer's limitation by designing lower-complexity alternatives such as the Longformer, Reformer, Linformer, and Performer. However, due to the wide range of solutions, it has become challenging for researchers and practitioners to determine which methods to apply in practice in order to meet the desired trade-off between capacity, computation, and memory. This survey addresses this issue by investigating popular approaches to make Transformers faster and lighter and by providing a comprehensive explanation of the methods' strengths, limitations, and underlying assumptions.
\end{abstract}

\ccsdesc[500]{Computing methodologies~Neural networks}

\keywords{Deep Learning, Efficient Transformer, Self-Attention, Survey}

\maketitle

\section{Introduction}
\label{sec:introduction}

\begin{figure}[htb]
  \centering
  \includegraphics[scale=0.47]{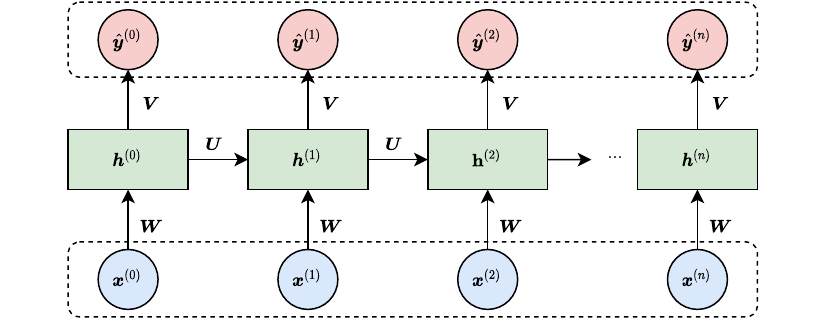}
  \Description[The computational graph of a recurrent neural network.]{The computational graph of a recurrent neural network.}
  \caption{The computational graph of a recurrent neural network. The input and output sequences are depicted in blue and red, respectively. The position, also known as the time-step, is indicated in superscript. The weight matrices $\boldsymbol{W}$, $\boldsymbol{U}$, and $\boldsymbol{V}$ are shared across all positions. Reproduced with permission~\citep{fournier21}. Copyright 2021 IEEE.}
  \label{fig:rnn}
\end{figure}

Sequences arise naturally in a wide range of domains, notably in natural language, biology, and software executions. \citet{10.5555/104339.104340} introduced a family of models called recurrent neural networks (RNNs) based on the idea of parameter sharing to process variable-length sequences. Given an input sequence $\boldsymbol{X}$ comprising $n$ tokens $\boldsymbol{x}^{(i)}$ of dimension $d$, recurrent neural networks iteratively construct a sequence of hidden representations $\boldsymbol{h}^{(i)}$ and produce a sequence of outputs $\boldsymbol{y}^{(i)}$  as illustrated in Figure~\ref{fig:rnn}. Unfortunately, vanilla RNNs often suffer from vanishing or exploding gradients, which prevent them from learning long-term dependencies. \citet{HochSchm97} addressed this limitation with the now widely popular long short-term memory (LSTM) network, which circumvents the gradient issues with paths through time. \citet{cho2014learning} later improved over the LSTM with the simpler gated recurrent unit (GRU).

\begin{figure}[htb]
  \centering
  \includegraphics[scale=0.47]{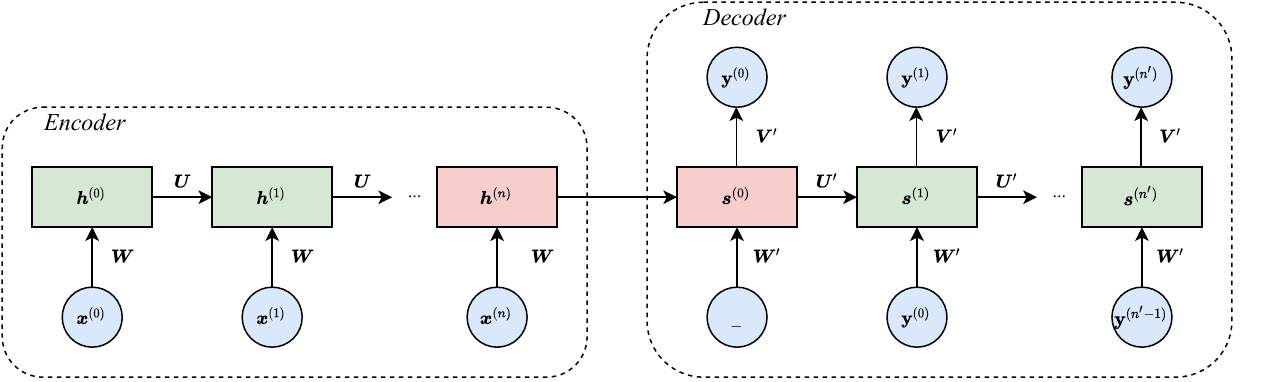}
  \Description[The sequence-to-sequence framework where the encoder and decoder are recurrent neural networks.]{The sequence-to-sequence framework where the encoder and decoder are recurrent neural networks.}
  \caption{The sequence-to-sequence framework where the encoder and decoder are recurrent neural networks. The input sequence (blue) is encoded into a fixed-size context $\boldsymbol{h}^{(n)}$ (red), which serves as the initial state of the decoder. Reproduced with permission~\citep{fournier21}. Copyright 2021 IEEE.}
  \label{fig:seq-to-seq}
\end{figure}

Recurrent neural networks align the input and output sequences, that is, there is a one-to-one mapping between the two sequences. Depending on the task, this property of RNNs may be too restrictive: for instance, translation requires outputting a sequence whose size is often different from that of the input while aligning tokens at different positions. \citet{NIPS2014_5346} addressed this limitation by introducing the sequence-to-sequence framework in which a first network (encoder) processes the entire input sequence and returns its last hidden representation $\boldsymbol{h}^{(n)}$, effectively encoding the input into a fixed-size vector called context. The context then serves as the initial state for a second network (decoder), which generates the output sequence in an autoregressive manner. The decoding stops when a special end-of-sequence token is generated. Figure~\ref{fig:seq-to-seq} illustrates the sequence-to-sequence framework.

In practice, the fixed-size nature of the hidden representation hinders the effectiveness of recurrent neural networks~\citep{DBLP:journals/corr/ChengDL16}.  Indeed, as the input sequence is processed, information is iteratively stored into the hidden representation that may be too small to retain all the relevant information for the task. In that case, useful data is inevitably lost, which may significantly impact the model's performance. \citet{Bahdanau2014} introduced an alignment mechanism called inter-attention to overcome the bottleneck of the sequence-to-sequence framework. This attention mechanism computes a different representation of the input for each output step, effectively allowing the decoder to ``look at'' the relevant part(s) of the input for each output step. Thereby, the inter-attention alleviates the encoder's burden to encode all information about the input sequence into a fixed-size vector. Formally, the context is the weighted sum of the encoder's hidden representations $\boldsymbol{h}_i $, for $i=1, \ldots, n$, where the weights are computed with a feed-forward neural network. For a comprehensive survey of the attention mechanism, we refer the reader to \citet{Galassi_2020} and \citet{weng2018attention}.  Figure~\ref{fig:attention} illustrates the inter-attention mechanism.

Moreover, recurrent neural networks do not scale efficiently to longer sequences due to their iterative nature~\citep{NIPS2017_7181}. In particular, RNNs struggle to learn dependencies between distant positions. One measure of this limitation is the relative effective context length (RECL) introduced by~\citet{dai-etal-2019-transformer}. The RECL is the largest context length that leads to a substantial relative gain over the best model. In other words, increasing the context length over the RECL yields a negligible increase in performance over the best model. The authors estimated that the relative effective context length of LSTMs on natural language data is limited to approximately 400 words. Besides, \citet{khandelwal2018sharp} empirically observed that LSTMs sharply model recent positions but only vaguely remember the distant past.

\subsection{Transformer}

\begin{figure}[!htb]
  \centering
  \includegraphics[scale=0.47]{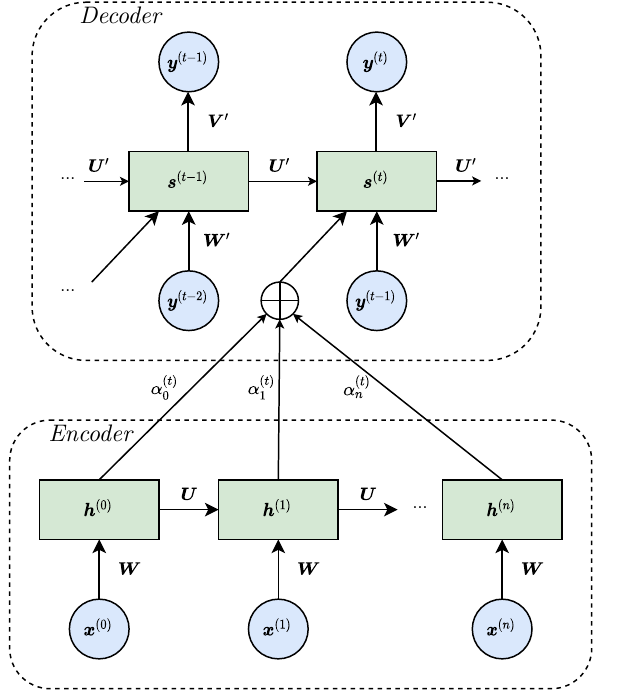}
  \Description[The inter-attention mechanism.]{The inter-attention mechanism.}
  \caption{The inter-attention mechanism. The attention weight $\alpha_i^{(t)}$ depicts the strength with which the $i$-th encoder hidden representation $h^{(i)}$ contributes to the context of $t$-th decoder step. Reproduced with permission~\citep{fournier21}. Copyright 2021 IEEE.}
  \label{fig:attention}
\end{figure}

This inherent limitation of recurrent neural networks has prevented them from being successfully applied to domains that require processing long sequences such as DNA. To overcome this limitation, \citet{NIPS2017_7181} introduced the \emph{Transformer}, a sequence-to-sequence model built without recurrences. Instead, the Transformer relies solely on the attention mechanism: the inter-attention between the encoder and decoder (see Figure~\ref{fig:attention}), and the self-attention, also known as intra-attention, within the encoder and decoder. The self-attention's main advantage is its ability to relate any two positions of the input sequence regardless of their distance, thus increasing performance significantly on a wide range of tasks, including natural language processing (NLP)~\citep{brown2020language, devlin2019bert, NIPS2017_7181}, computer vision~\citep{carion2020endtoend, dosovitskiy2020image, 10.1145/3505244}, speech recognition~\citep{gulati2020conformer, 9413901,Zhang2020TransformerTA}, and biological sequence analysis~\citep{NEURIPS2020_c8512d14}. \citet{karita2019a} evaluated a Transformer against a sequence-to-sequence Bi-LSTM baseline on automatic speech recognition (ASR), speech translation (ST), and text-to-speech (TTS). The attention-based models outperformed the baseline on 13 corpora out of 15 for monolingual ASR and realized more than 10\% relative improvement in 8 languages out of 10 for multilingual ASR. The Transformer improved the BLEU score from 16.5 for the baseline to 17.2 on ST while performing on par for TTS. Table~\ref{tab:comparison} reports the performance improvements brought by popular Transformer architectures over previous state-of-the-art models across different domains. As of this paper's writing, the Transformer has become the de facto model for numerous sequence processing tasks.

\begin{table}[htb]
  \centering
  \caption{Relative improvements brought by popular Transformer architectures over previous state-of-the-art models. Absolute differences are reported between parenthesis. Sources are: \citep{NIPS2017_7181} for machine translation, \citep{dosovitskiy2020image} for image classification, \citep{devlin2019bert, Radford2018ImprovingLU} for text classification, and \citep{liu2020improving} for speech-to-text.}
  \resizebox{\textwidth}{!}{%
  \begin{tabular}{lllll}
    Task                                  & Dataset                  & Previous SOTA                                               & Transformer's Architecture             & Relative Improvement                       \\ \hline
    \multirow{2}{*}{Machine Translation}  & newstest2014 (EN-to-DE)  & MoE (GNMT)~\citep{shazeer2017outrageously}                  & Vanilla~\citep{NIPS2017_7181}          & 9.1\% (+2.37 BLEU\protect\footnotemark[3]) \\
                                          & newstest2014 (EN-to-FR)  &                                                             &                                        & 3.1\% (+1.24 BLEU)                         \\ \hline
    \multirow{4}{*}{Image Classification} & ImageNet                 & Noisy Student (EfficientNet-L2)~\citep{xie2020selftraining} & ViT~\citep{dosovitskiy2020image}       & 0.2\% (+0.15\% Acc)                        \\
                                          & CIFAR-10                 & BiT-L (ResNet152x4)~\citep{kolesnikov2020big}               &                                        & 0.1\% (+0.13\% Acc)                        \\
                                          & CIFAR-100                &                                                             &                                        & 1.1\% (+1.04\% Acc)                        \\
                                          & VTAB (19 tasks)          &                                                             &                                        & 1.8\% (+1.34\% Acc)                        \\ \hline
    \multirow{2}{*}{Text Classification}  & SST2                     & Sparse byte mLSTM~\citep{Gray2017GPUKF}                     & BERT\citep{devlin2019bert}             & 1.8\% (+1.70\% Acc)                        \\
                                          & CoLA                     & Single-task BiLSTM + ELMo + Attn~\cite{wang2019glue}        &                                        & 72.9\% (+25.5 MC\protect\footnotemark[4])  \\ \hline
    \multirow{2}{*}{Speech-to-text}       & librispeech (test-clean) & LAS (LSTM)~\citep{chan2015listen, park2019specaugment}      & Convformer~\citep{gulati2020conformer} & 13.6\% (-0.3 WER\protect\footnotemark[5])  \\
                                          & librispeech (test-other) &                                                             &                                        & 25.0\% (-1.3 WER)                          \\ \hline
  \end{tabular}%
  }
  \label{tab:comparison}
\end{table}

As an illustration of an end-to-end application of the Transformer, let us consider the speech recognition task. In hybrid approaches, the recognition system consists of independently trained machine learning components, often an acoustic model, a pronunciation model, and a language model. Instead, in end-to-end approaches, the recognition system consists of a single model comprising several parts trained together. \citet{Zhang2020TransformerTA} introduced an end-to-end speech recognition model based on Transformer encoders called the Transformer Transducer that outperformed previous hybrid and end-to-end approaches on the LibriSpeech benchmarks.

The Transformer's capacity comes at the cost of a quadratic computational and memory complexity with respect to the sequence length. Therefore, training large Transformers is prohibitively slow and expensive. For instance, \citet{liu2019roberta} introduced RoBERTa, which was pre-trained on 1024 high-end V100 graphics processing units (GPUs) for approximately a day. Although numerous large pre-trained Transformers have been publicly released, fine-tuning them on the tasks of interest is still computationally expensive. Furthermore, the sequence lengths are restricted by the amount of memory available. Indeed, practitioners typically use large mini-batches with relatively short sequences because the Transformer's optimization is known to be particularly unstable with small mini-batches. Typically, a GPU with 16 GB of memory handles sequences up to 512 words. Consequently, there exists an actual need for lighter and faster Transformers as only a few large organizations can afford to train massive models. As of the writing of this paper, the largest dense Transformer is GPT-3~\citep{brown2020language} which requires 355 years to train on a V100 GPU, costing around 4,600,000\$ of cloud instances\footnote[1]{\url{https://lambdalabs.com/blog/demystifying-gpt-3}}.
\footnotetext[2]{Bilingual evaluation understudy (BLEU), higher is better.}
\footnotetext[3]{Matthews correlation (MC) coefficient, higher is better.}
\footnotetext[4]{Word error rate (WER), lower is better.}
\setcounter{footnote}{4}

\subsection{Lighter and Faster Transformers}

Over the years, numerous approaches have been proposed to reduce the computational and memory costs
of neural networks, many of which have been applied to Transformers. In this paper, such methods are referred to as \emph{general} since they apply, and have been applied, to a wide range of models. General methods are often orthogonal, and consequently, several of them may be combined to precisely fine-tune the network's capacity, computational cost, and memory usage. However, general methods may be insufficient as the model complexity typically remains unchanged. Therefore, many works introduced lower-complexity variations of the Transformer, referred to as \texttt{x}-formers.  In this survey, the Transformer's alternatives are categorized depending on whether they sparsify the attention, factorize it, or modify the network's architecture. Please note that this survey aims to provide a comprehensive summary of the methods that improve the Transformer's efficiency and that fine-grained taxonomies have already been proposed by \citet{tay2020efficient} and \citet{DBLP:journals/corr/abs-2106-04554}. Accordingly, our taxonomy will remain purposefully coarse. 

Recently, \citet{tolstikhin2021mlpmixer} and \citet{liu2021pay} amongst others argued that the powerful yet expensive self-attention mechanism is not necessary to achieve state-of-the-art results and thus challenged the preconception that the self-attention is the source of the Transformer's success. Consequently, they introduced networks without self-attention that are competitive with Transformers for image classification and language modelling at the same computational cost. \citet{metaformer} expanded on this idea with a more general and flexible architecture called MetaFormer where the mechanism to relate the tokens is not specified while the other components are kept the same as the Transformer. Despite the recent success of attention-free architectures, such networks are outside the scope of this paper as they arguably remove the Transformer's core mechanism and are discussed in appendix.

The remainder of this survey is organized as follows. Section~\ref{sec:attention} introduces the Transformer's architecture and the origin of the quadratic complexity. Section~\ref{sec:generic} investigates the popular general methods that have been applied to Transformers to reduce the computations and memory footprint. Section~\ref{sec:specialized} explores the recent lower-complexity Transformers. Section~\ref{sec:shortcomings} explains the limitations of the different approaches and the current evaluation methodology, Section~\ref{sec:impact} provides a discussion on the broader impact of lighter and faster Transformers, and Section~\ref{sec:directions} points out potential future research directions. Finally, Section~\ref{sec:conclusion} concludes this survey. Practitioners and researchers can find detailed practical guidelines regarding the general and specialized methods in appendix, as well as a summary of the specialized methods (see Table~\ref{tab:specialized_methods}) and a discussion about some of the most popular attention-free alternatives.

\section{Transformer}
\label{sec:attention}

  This section formally introduces the attention mechanism, the Transformer's architecture, and the root cause of its quadratic complexity.

  \begin{figure}[htb]
    \centering
    \includegraphics[scale=0.47]{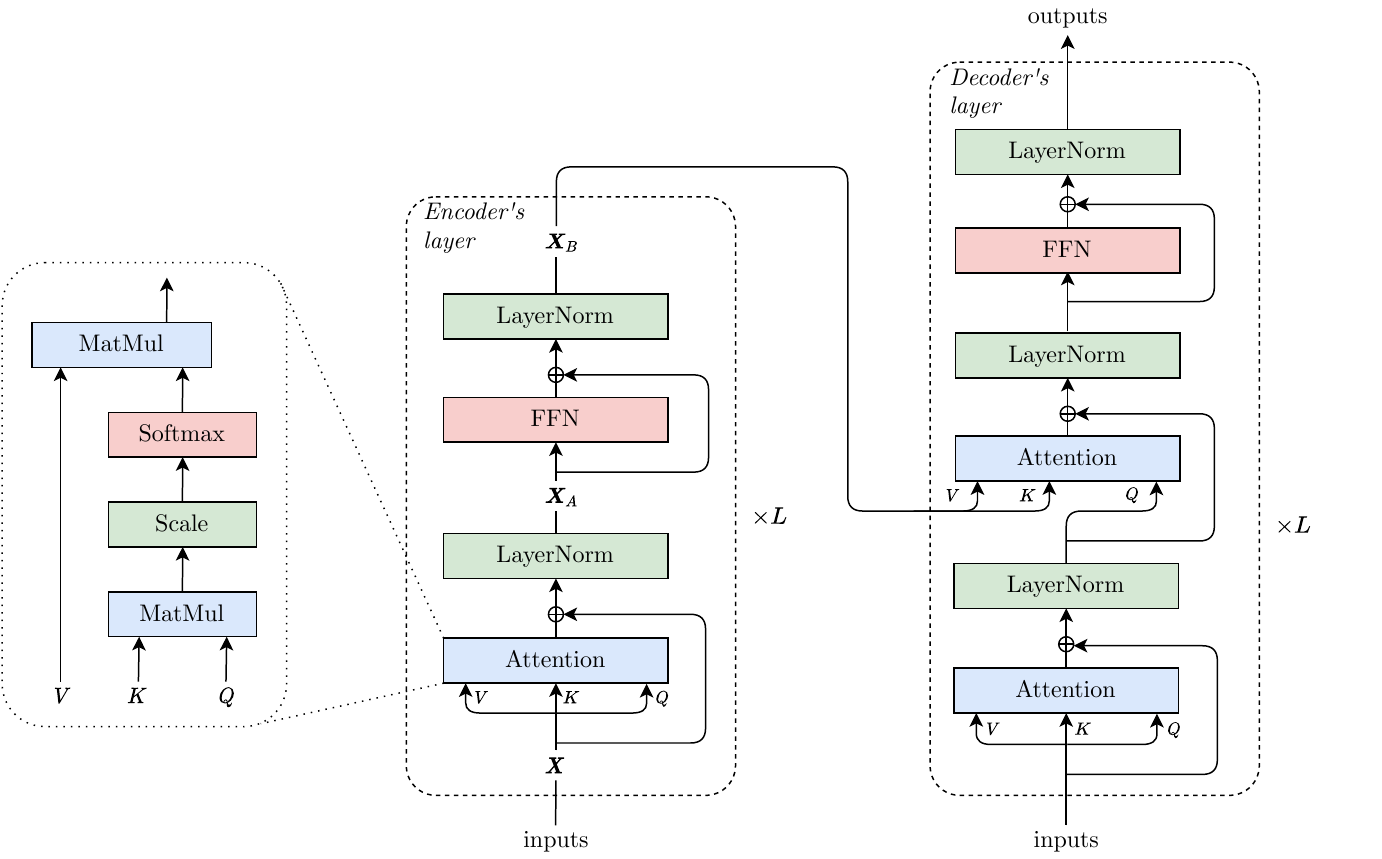}
    \Description[The Transformer's computational graph.]{The Transformer's computational graph.}
    \caption{The Transformer's computational graph~\citep{NIPS2017_7181}. From left to right, the scaled dot product self-attention, the encoder, and the decoder. Note that both the encoder and decoder comprise $L$ identical layers, of which only one is depicted.}
    \label{fig:transformer}
  \end{figure}

  \subsection{Attention Mechanism}

  The attention mechanism relies on three matrices, namely $\boldsymbol{Q}, \boldsymbol{K}, \boldsymbol{V} \in \mathbb{R}^{n \times d}$, commonly referred to as ``queries'', ``keys'', and ``values'', respectively. The attention outputs the sum of the values weighted by a compatibility or alignment score between each token, which is computed with the function $\mathrm{Score}(\boldsymbol{Q}, \boldsymbol{K}) \in \mathbb{R}^{n \times n}$. Intuitively, if the $i$-th query is highly compatible with the $j$-th key, then the $j$-th value greatly contributes to the $i$-th attention's output. The attention mechanism may be written as:
  \begin{equation}
    \mathrm{Attention}(\boldsymbol{Q}, \boldsymbol{K}, \boldsymbol{V}) = \mathrm{Score}(\boldsymbol{Q}, \boldsymbol{K})\boldsymbol{V}.
    \label{eq:attn_score}
  \end{equation}

  Since the compatibility score directly controls the alignment between the tokens, many functions have been proposed. In the original paper, the Transformer relies on the \emph{scaled dot product attention}. The \emph{dot product} refers to the computation of the compatibility score between a single query and a single key. In practice, however, the compatibility scores are computed simultaneously for every query and key by multiplying $\boldsymbol{Q}$ with $\boldsymbol{K}^\top$. Indeed, the $(i, j)$ entry of the $\boldsymbol{QK}^\top$ multiplication is equal to the dot product between the $i$-th query and the $j$-th key. In order to obtain a probability distribution over the positions, referred to as attention weights, each row of $\boldsymbol{QK}^\top$ is passed through a $\mathrm{Softmax}$ function defined as follows:
  \begin{equation}
    \mathrm{Softmax}(\boldsymbol{x})_i=\frac{e^{x_i}}{\sum_{j=1}^n e^{x_j}} \quad \mathrm{for\ } i=1, \dots, n.
  \end{equation}
  \noindent where $\boldsymbol{x}\in\mathbb{R}^n$. Since the dot product grows large in magnitude for large values of $d$, thereby pushing the $\mathrm{Softmax}$ into a region of small gradients, a \emph{scaling} factor $\sqrt{d}$ is introduced. Thus, the scaled dot product attention is given by:
  \begin{equation}
    \mathrm{Attention}(\boldsymbol{Q}, \boldsymbol{K}, \boldsymbol{V}) = \mathrm{Softmax}\left(\frac{\boldsymbol{QK}^\top}{\sqrt{d}}\right)\boldsymbol{V}.
    \label{eq:attention}
  \end{equation}
  
  Nonetheless, the attention presented above may not be flexible enough if the relevant information for the task is scattered across different regions of the input space. That is due in part to the $\mathrm{Softmax}$ being exponential, which amplifies the differences between the values. As a result, only a few attention weights are large, i.e., only a few positions are strongly attended. \citet{NIPS2017_7181} addressed this limitation with the multi-head attention. The $d$-dimensional queries, keys and values matrices are first linearly projected $h$ times with distinct, learned projections to $d_k$, $d_k$ and $d_v$ dimensions, respectively. On each projection, an independent attention instance called \emph{head} is applied, and the output of each attention head is concatenated before being linearly projected. The Transformer's multi-head scaled dot product attention is given by:
  \begin{gather}
    \mathrm{MultiHead}(\boldsymbol{Q}, \boldsymbol{K}, \boldsymbol{V}) = [\mathrm{head}_1;...;\mathrm{head}_h]\boldsymbol{W}^O.\\
    \mathrm{head}_i =\mathrm{Softmax}\left(\frac{\boldsymbol{QW}^Q_i (\boldsymbol{KW}^K_i)^\top}{\sqrt{d_k}}\right)\boldsymbol{VW}^V_i.
  \end{gather}
  \noindent where $\boldsymbol{W}^Q_i \in \mathbb{R}^{d \times d_k}$, $\boldsymbol{W}^K_i \in \mathbb{R}^{d \times d_k}$, $\boldsymbol{W}^V_i \in \mathbb{R}^{d \times d_v}$ are the matrices that project the queries, keys, and values into the $i$-th subspace, respectively, and where $\boldsymbol{W}^O \in \mathbb{R}^{hd_v \times d}$ is the matrix that computes a linear transformation of the heads. Typically, $d_k=d/h$ where $d$ is the input and output dimension, and $h$ is the number of heads. For the sake of clarity, methods that modify the attention will be explained in the context of a single head (see Equation~\ref{eq:attention}).

  Thus far, the attention mechanism has been described as a general method. The Transformer relies on two specific instances of this mechanism: the intra-attention, popularly known as self-attention, and the inter-attention, sometimes referred to as cross-attention. In the case of inter-attention, the queries correspond to the decoder's hidden representations, and the keys and values are the encoder's outputs. It allows the decoder to look at the relevant parts of the input to produce the output. In the case of self-attention, the three matrices are linear projections of the layer's input, which allows the encoder and decoder to focus on the relevant part of the sequence for each position, similarly to the inter-attention depicted in Figure~\ref{fig:attention}.

  \subsection{Encoder}

  The Transformer's encoder is a function defined as the composition of $L$ identical layers or blocks, each composed of two sub-layers. The first sub-layer is the aforementioned self-attention mechanism. The second sub-layer is a simple fully connected feed-forward network applied position-wise, that is, independently and identically to every position. The feed-forward network increases the encoder's expressiveness and transforms the self-attention's output for the next layer.

  Inspired by ResNet~\citep{he2015deep}, a skip connection, shortcut connection, or residual connection is applied around each sub-layer to create a direct path for the gradient to flow throughout the network. Notably, residual connections make the training of very deep neural networks more stable. Additionally, both sub-layers' outputs are normalized after the residual connection with the layer normalization technique, referred to as $\mathrm{LayerNorm}$ \cite{ba2016layer}. Normalization is a widely adopted technique in deep learning that enables faster and more stable training. Although the rationale behind the normalization's empirical success is not yet fully understood~\citep{li2020reconciling}, it has been conjectured that this results from a smoother optimization landscape, and to a lesser extent, from a reduction in internal covariance shift~\citep{santurkar2019does}. Figure~\ref{fig:transformer} depicts the computational graph of an encoder's layer.

  In natural language processing, the input sequence $\boldsymbol{X}$ would typically represent a sentence or a paragraph, and the token $\boldsymbol{x}^{(i)}$ would be its $i$-th word or subword embedding. Each encoder's layer is  given by:
  \begin{gather}
    \boldsymbol{X}_A = \mathrm{LayerNorm}(\mathrm{Attention}(\boldsymbol{Q}, \boldsymbol{K}, \boldsymbol{V}) + \boldsymbol{X})\label{eq:1}\\
    \boldsymbol{X}_B = \mathrm{LayerNorm}(\mathrm{FFN}(\boldsymbol{X}_A) + \boldsymbol{X}_A)
  \end{gather}
  where $\boldsymbol{X}$ and $\boldsymbol{X}_B$ are the layer's input and output, respectively, and $\boldsymbol{Q}$, $\boldsymbol{K}$, and $\boldsymbol{V}$ are linear projections of $\boldsymbol{X}$.

  The feed-forward network is given by:
  \begin{equation}
    \mathrm{FFN}(\boldsymbol{x})=\max(0, \boldsymbol{x}\boldsymbol{W}_1+\boldsymbol{b}_1)\boldsymbol{W}_2 + \boldsymbol{b}_2
  \end{equation}
  where $\boldsymbol{W}_1 \in \mathbb{R}^{d \times d_f}$ and $\boldsymbol{W}_2 \in \mathbb{R}^{d_f \times d}$, and where $d_f$ is the dimension of the hidden layer. Note that the feed-forward network is defined for a row vector since it is applied position-wise, that is, it is independently and identically applied to every position or row.

  Finally, the position-wise layer normalization is given by:
  \begin{equation}
    \mathrm{LayerNorm}(\boldsymbol{x})=\boldsymbol{g}\odot \frac{\boldsymbol{x}-\mu}{\sqrt{\sigma^2+\epsilon}} + \boldsymbol{b}
  \end{equation}
  \noindent where $\odot$ denotes the element-wise (Hadamard) product, where the average $\mu$ and the standard deviation $\sigma$ are computed from all of the summed inputs, where the gain $\boldsymbol{g}$ and the bias $\boldsymbol{b}$ are learned parameters of dimension $d$, and where $\epsilon$ is a small constant used in practice for numerical stability.

  \subsection{Decoder}

  The decoder is also composed of $L$ identical layers. Although it is common for the decoder to have the same number of layers as the encoder, one may adjust their depth independently. Each decoder's layer comprises three sub-layers. The first sub-layer is the self-attention mechanism, as in the encoder, except that future positions are masked. Indeed, while the encoder is allowed to look at future positions since the input sequence is entirely available, the decoder is autoregressive and thus cannot look at future positions since they have not yet been predicted. Therefore, the $i$-th position may only attend to positions less than $i$. The second sub-layer is the inter-attention mechanism, which helps the decoder focus on the relevant parts of the input. Finally, the third sub-layer is a simple feed-forward network. As for the encoder, a residual connection and a layer normalization are applied to each sub-layer.

Note that the decoder may be safely omitted when the task does not require the sequence-to-sequence framework, such as sentiment analysis, which predicts whether a sentence is positive. One of the most popular encoder-only Transformers is the Bidirectional Encoder Representations from Transformers  (BERT)~\citep{devlin2019bert}, a state-of-the-art language model that learns contextualized embeddings. Nonetheless, autoregressive tasks such as machine translation still require the sequence-to-sequence framework.

\subsection{Complexity}
\label{ssec:complexity}

\begin{figure}[htb]
  \centering
  \includegraphics[scale=0.2]{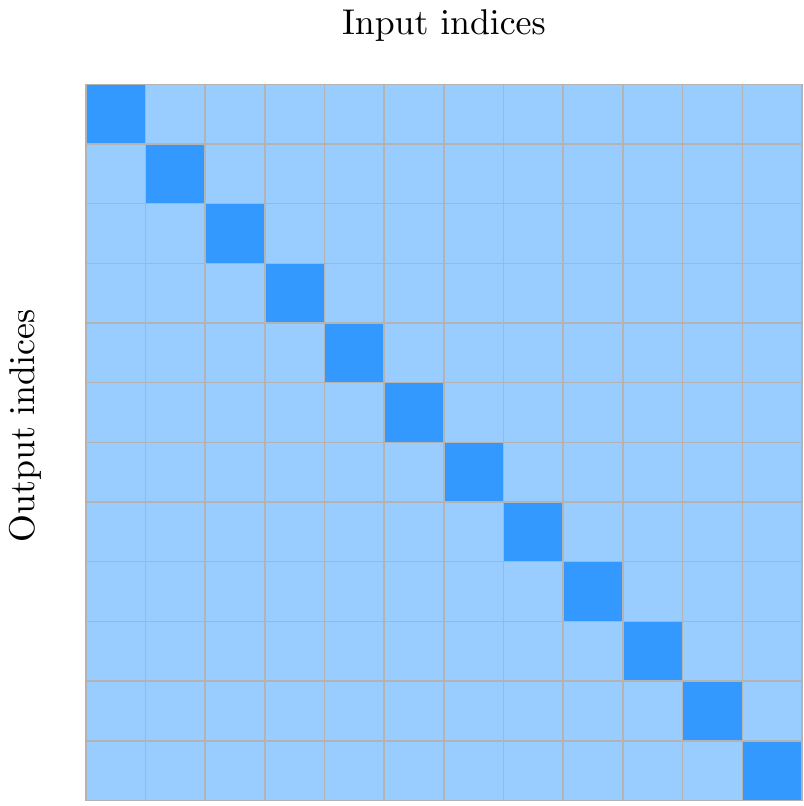}
  \Description[The connectivity matrix of the full attention.]{The connectivity matrix of the full attention.}
  \caption{The connectivity matrix of the full attention. The $i$-th output position attends to the $j$-th input position if, and only if, the cell $(i, j)$ is coloured. The diagonal is highlighted to ease the reading.}
  \label{fig:full}
\end{figure}

Intuitively, the quadratic complexity emerges from the computation of the compatibility score between every pair of positions. More precisely, the $\boldsymbol{QK}^\top$ multiplication requires $n^2$ computations and memory. Such attention is said to be full since any output position is able to attend to any input position. The attention pattern is visualized by means of a connectivity matrix, which indicates the input positions that each output position is able to attend (see Figure~\ref{fig:full}).

What justifies such efforts from the community to improve the Transformer's efficiency? In our opinion, there are three primary motivations: affordability, scalability, and ecology.

The foremost reason is affordability. The Transformer has largely surpassed convolutional and recurrent neural networks and achieved new state-of-the-art results across many tasks. However, those networks have a linear complexity with respect to the sequence length~\citep{NIPS2017_7181}, making them affordable to most researchers and practitioners. As explained by \citet{Strubell_Ganesh_McCallum_2020}, this creates three major issues: (1) it stifles creativity as researchers and practitioners that do not have access to considerable resources are not able to experiment with Transformers, (2) it reinforces the ``rich get richer'' cycle where successful labs and companies receive more funding due to their existing accomplishments with Transformers, and (3) it forces smaller labs and companies to rely on private cloud services that end up more expensive. 

The second reason is scalability. The quadratic complexity prevents researchers and practitioners, even those with access to considerable resources, from applying Transformers on long sequences such as entire chapters or books, high-resolution images or videos, and DNA. 

The third reason is ecology. It is now more apparent than ever that we must cut carbon dioxide (CO2) emissions in half over the next decade to limit global warming. The large-scale infrastructures used by the deep learning community consume a considerable amount of electricity, which is mainly produced by non-renewable sources such as coal or gas~\citep{iea}.

Thereby, the following sections investigate popular and novel methods to make Transformers faster and lighter. 

\section{General Approaches}
\label{sec:generic}

Computational resources have always been a limiting factor for deep learning models~\citep{NIPS1989_250}. Therefore, numerous approaches have been proposed throughout the years to design faster and lighter models. This section introduces the most popular techniques that apply to virtually all neural networks.

\textbf{Gradient Checkpointing}~\citep{chen2016training}: Intermediate results computed during the forward pass, also referred to as activations, are required to compute the gradients during the backward pass; therefore, they are stored in memory. Activations typically account for most of the memory during training: given an $l$-layer network, the number of intermediate results is proportional to the number of layers ($\mathcal{O}(l)$). With gradient checkpointing, also known as rematerialization, activations are stored only for a subset of the layers. However, they must be recomputed during the backward pass, trading memory for computations. In the extreme case where no activations are stored, the memory usage becomes constant ($\mathcal{O}(1)$) at the cost of a quadratic number of computations with respect to the number of layers ($\mathcal{O}(l^2)$). \citet{chen2016training} designed a scheme to select the preserved values that reduces the memory requirement from $\mathcal{O}(l)$ to $\mathcal{O}(\sqrt{l})$ at the cost of a single additional forward pass per mini-batch. OpenAI implementation of gradient checkpointing~\citep{openai_checkpointing} obtains an impressive $10\times$ reduction in memory at the cost of a 20\% increase in computation time.

\textbf{Reversible Layers}~\citep{DBLP:journals/corr/DinhKB14, gomez2017reversible, DBLP:conf/iclr/DinhSB17}: As explained above, the back-propagation requires the activations of all intermediate layers, which are either stored in memory during the forward pass or recomputed during the backward pass. As a solution to the latter case, reversible layers allow their activation to be reconstructed exactly from the next layer; therefore, activations must only be stored for one layer and their memory cost becomes independent of the network's depth. More formally, each reversible layer takes as input $(x_1, x_2)$ and outputs $(y_1, y_2)$ such that $ y_1 = x_1 + f(x_2)$ and $y_2 = x_2 + g(y_1)$. Each layer's activations are easily reconstructed as $x_2 = y_2 - g(y_1)$ and $x_1 = y_1 - f(x_2)$.

\citet{Kitaev2020Reformer} used reversible layers in their Transformer, called the Reformer, by combining the attention and feed-forward sub-layers inside a reversible layer. Specifically, $f(.)$ and $g(.)$ were the $\mathrm{Attention(.)}$ and $\mathrm{FFN(.)}$ functions, respectively. The authors observed that reversible layers reduced the memory usage of a 3-layer Transformer without degrading its performance. Nonetheless, reversible layers add numerical errors that accumulate over multiple layers and may degrade the model performance. Therefore, they are not suited for very deep networks.

Gradient checkpointing and reversible layers are very much alike in that they trade computations for memory by recomputing activations during backpropagation. This trade-off is sometimes necessary: although computation bottlenecks entail longer running times, memory bottlenecks are critical as they prevent using the model altogether.

\textbf{Parameter Sharing}: A simple approach to reduce the number of trainable parameters is to impose sets of parameters to be equal in different parts of the network. In other words, the same parameters are used for multiple operations but need to be stored only once in memory. Such a technique is often referred to as parameter sharing, weight tying, or weight replication. As explained in Section~\ref{sec:introduction} and illustrated in Figure~\ref{fig:rnn}, recurrent neural networks are built around this idea of parameter sharing to process variable-length sequences. Parameter sharing has also been applied to Transformers. For instance, the Linformer~\citep{2020arXiv200604768W} shares projection matrices across heads and layers, and the Reformer~\citep{Kitaev2020Reformer} shares its queries and keys parameters, that is, $\boldsymbol{W}^Q=\boldsymbol{W}^K$. Both authors investigated the impact of parameter sharing and concluded that it did not degrade their respective models' performance on their tasks. \citet{lan2020albert} shared all parameters between layers, which drastically reduced the number of parameters but also decreased the performance by up to 2.5\% on average. They observed that sharing only the attention parameters resulted in a slight drop in performance of 0.7\% on average. The decrease in performance is to be expected since parameter sharing reduces the number of free parameters, hence the model's capacity.

\textbf{Pruning}~\citep{NIPS1989_250}: Smaller neural networks are not only faster and lighter, but they are also more likely to generalize better than larger models because they presumably extract underlying explanatory factors without redundancy. To reduce the model size, weights with a small saliency, that is, whose deletion have a small effect on the loss, may be removed from large models after training. Methods that consider individual weights are said to be \emph{unstructured}, and methods that consider pieces of the network structure such as attention heads or layers are said to be \emph{structured}. Many structured and unstructured pruning schemes have been proposed, several of which have been applied to Transformers. For instance, \citet{2020arXiv200403844S} reduced the size of BERT by 40\% by dropping complete layers while retaining between 97 and 98\% of its original performance, and \citet{michel2019sixteen} pruned away between 20\% and 40\% of BERT attention heads without any significant loss in performance. Recently, the lottery ticket hypothesis has brought a new justification to pruning neural networks. As introduced by \citet{frankle2019lottery}, the hypothesis states that a ``\textit{randomly-initialized, dense neural network contains a subnetwork that is initialized such that -- when trained in isolation -- it can match the test accuracy of the original network after training for at most the same number of iterations.}''. \citet{prasanna2020bert} successfully verified this hypothesis on BERT, even noticing that BERT worst subnetworks remain highly trainable. Nonetheless, pruning has two limitations: a large model must be trained, and unstructured pruning schemes produce sparse models unoptimized for modern GPUs and tensor processing units (TPUs).

\textbf{Knowledge Distillation}~\citep{ba2014deep, hinton2015distilling}: The knowledge of a large model or an ensemble of models (teacher) is transferred to a single smaller model (student) by training the student to reproduce the teacher's outputs or its internal behaviour. The cumbersome teacher is then discarded, and the student is used at inference time. Given a parameter budget, networks trained with knowledge distillation usually outperform models directly trained on the task. \citet{sanh2020distilbert}, \citet{tsai2019small}, and \citet{jiao2020tinybert} applied different knowledge distillation schemes on the original BERT~\citep{devlin2019bert} to obtain lighter and faster models called DistilBERT, MiniBERT, and TinyBERT, respectively. Table~\ref{tab:distil} reports their compression, speed-up, and performance. Although knowledge distillation achieves impressive compression ratios and performance trade-offs, a large teacher model still needs to be trained, and the student may perform significantly worse than the teacher. For instance, $\mathrm{BERT}_{\mathrm{BASE}}$ achieves an accuracy of 52.8\% on the CoLA task~\citep{warstadt2019neural}, while DistilBERT and TinyBERT only achieve 32.8\% and 44.1\%, respectively, according to \citet{jiao2020tinybert}.

\begin{table}[htb]
  \centering
  \small
  \caption{Multiple knowledge distillations of $\mathrm{BERT}_{\mathrm{BASE}}$. Speed-ups are evaluated on GPUs.}
  \label{tab:distil}
  \begin{tabular}{lccc}
    Model                                                  & Compression & Speed-up        & Mean Relative Performance\\ \hline
    $\mathrm{BERT}_{\mathrm{BASE}}$~\citep{devlin2019bert} & $1.0\times$ & $1.0\times$     & $100\%$                   \\ \hline
    DistilBERT~\citep{sanh2020distilbert}                  & $1.7\times$ & $1.6\times$     & $97\%$                    \\
    MiniBERT~\citep{tsai2019small}                         & $6.0\times$ & $2.6-4.3\times$ & $97-99\%$                 \\
    TinyBERT~\citep{jiao2020tinybert}                      & $7.5\times$ & $9.4\times$     & $97\%$                    \\ \hline
  \end{tabular}
\end{table}

\textbf{Mixed-Precision}~\citep{micikevicius2018mixed}: Modern GPUs and TPUs perform at least twice as many half-precision (16 bits) float operations as single-precision (32 bits) ones. A popular approach to accelerate training and reduce memory consumption is storing and computing the weights, activations, and gradients in half-precision. A master copy of the weights is stored in single-precision for numerical stability and minimal performance loss. Thanks to NVIDIA's Automatic Mixed-Precision included in some of the most popular deep learning libraries, namely TensorFlow, PyTorch, and MXNet, using mixed precision can be as simple as adding one line of code. Consequently, we highly recommend mixed-precision. \citet{jacob2017quantization} improved over this approach by quantizing both weights and activations as 8-bit integers and biases as 32-bit integers, effectively allowing inference to be performed using integer-only arithmetic.  Given a parameter matrix $\boldsymbol{W}$, $N$-bit quantization rounds each parameter to one of $2^N$ codewords corresponding to bins evenly spaced by a scale factor $s$ and shifted by a bias $z$ computed as follows:
    \begin{equation}
      s=\frac{\max\boldsymbol{W}-\min\boldsymbol{W}}{2^N - 1} \quad \mathrm{and} \quad z = \mathrm{round}\left(\frac{\min\boldsymbol{W}}{s}\right)
    \end{equation}
    Each parameter $W_{i, j}$ is quantized to its nearest codeword, and dequantized as:
    \begin{equation}
      \hat{W}_{i, j} = \left( \mathrm{round} \left( \frac{W_{i, j}}{s} + z \right) - z \right) \times s\\
    \end{equation}
 
In order to mitigate the performance loss associated with the low-precision approximation, Quantization Aware Training (QAT)~\citep{jacob2017quantization} quantizes the parameters during training. Since quantization is not differentiable, gradients are approximated with a straight-through approximator~\citep{bengio2013estimating}. Notably, \citet{zafrir2019q8bert} quantized all matrix product operations in BERT fully connected and embedding layers during training, reducing the memory footprint by 4$\times$ while retaining 99\% of the original accuracy on the GLUE~\citep{wang2019glue} and SQuAD~\citep{rajpurkar2016squad} tasks. \citet{fan2021training} achieved an even higher compression ratio with iterative product quantization (iPQ), which replaces vectors of weights by their assigned centroid, and quantization of those centroids. The authors reduced the size of a 16-layer Transformer by 25$\times$, making the model only 14 MB, while retaining 87\% of the original performance on the Wikitext-103~\citep{merity2016pointer} benchmark.

While pruning and knowledge distillation achieve faster and lighter models by reducing the number of parameters, mixed-precision and quantization instead reduce the number of bits per parameter.

\begin{figure}[htb]
  \centering
  \includegraphics[scale=0.47]{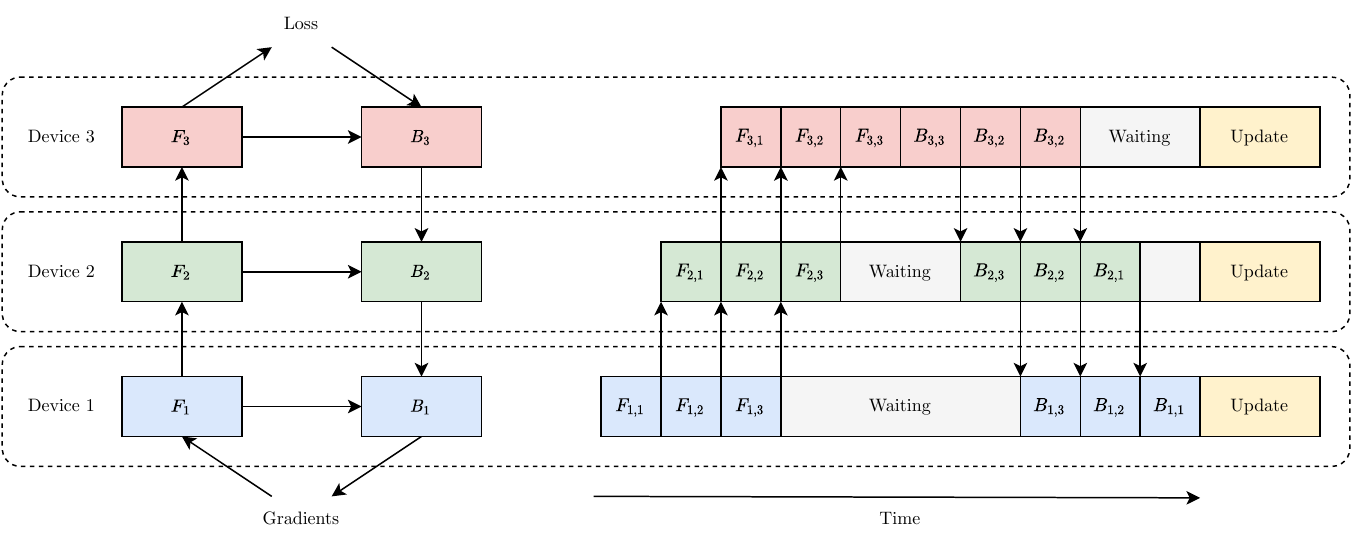}
  \Description[Micro-Batching applied to a model distributed across three devices.]{Micro-Batching applied to a model distributed across three devices.}
  \caption{Micro-Batching applied to a model distributed across three devices~\citep{huang2019gpipe}. $F_i$ and $B_i$ denotes the sequential forward and backward operations, respectively, performed by the $i$-th device. Computation on a device may start as soon as the previous device in the computational graph has processed the first micro-batch. Therefore, micro-batching reduces the waiting time of each device at the cost of inter-device communications. Note that the model update is done synchronously at the end.}
  \label{fig:gpipe}
\end{figure}

\textbf{Micro-Batching}~\citep{huang2019gpipe}: Increasing model capacity and data throughput are efficient strategies for improving performances in deep learning. However, increasing data throughput requires transferring large mini-batches to the accelerators' memory\footnote{An accelerator denotes any device that accelerates computation, such as a graphics or tensor processing unit.}, which is also used to store the model. One way to partially avoid the trade-off between mini-batch size and model size is to use model parallelism. GPipe~\citep{huang2019gpipe} is a model parallelism library that enables users to distribute a model by grouping layers into cells assigned to accelerators. To avoid the communication bottleneck between accelerators due to the forward and backward operations, the authors proposed a novel batch-splitting algorithm that further splits the mini-batch into micro-batches. As soon as the first accelerator finishes the forward operation of the layers assigned to it for a micro-batch, it sends the result over the communication link and starts processing the next micro-batch. After finishing the last micro-batch's forward operation, the accelerators wait for the first micro-batch's backwards operation results. This waiting time can be used to recompute the forward operation and further reduce memory usage, known as rematerialization. Finally, once the backward operation is completed on the last micro-batch, the algorithm sums all micro-batch's gradients to obtain the mini-batch's gradient (see Figure~\ref{fig:gpipe}). However, the result is not exact with layers that compute statistics across all mini-batch examples, such as a batch normalization layer~\citep{ioffe2015batch}. Finally, GPipe is compatible with data parallelism, where multiple mini-batches are processed in parallel.

\citet{huang2019gpipe} empirically demonstrated that GPipe allows the maximum Transformer size to scale linearly with the number of accelerators. For instance, a TPU v3 with 16Gb of memory can only fit a 3-layer Transformer. With GPipe, the same TPU is able to fit 13 layers, while 128 TPUs are able to fit 1663 layers, which is 127.9$\times$ more. Additionally, the authors distributed a 48-layer Transformer across 8 TPUs and reported that the training throughput was 4.8 times higher with 32 micro-batches than with a single one.

\begin{figure}[htb]
  \centering
  \includegraphics[scale=0.47]{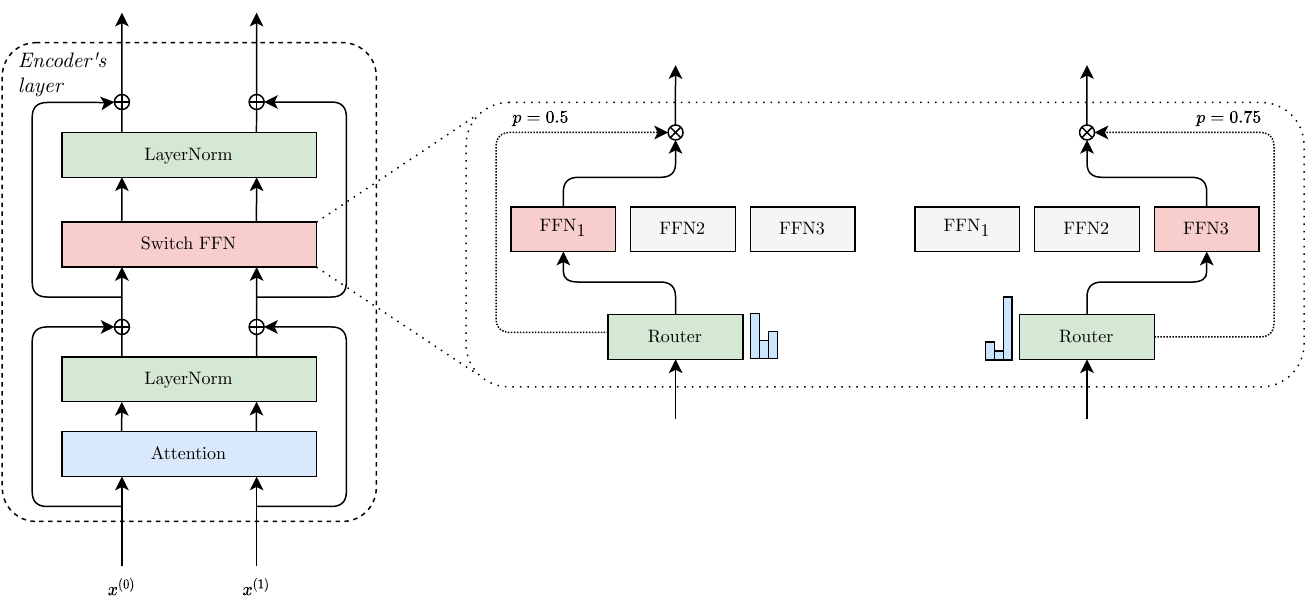}
  \Description[The computational graph of a single layer of the Switch Transformer's encoder.]{The computational graph of a single layer of the Switch Transformer's encoder.}
  \caption{The computational graph of a single layer of the Switch Transformer's encoder~\citep{fedus2021switch}. The Transformer's feed-forward network (FFN) has been replaced by a Switch FFN which independently routes each position to an expert. The expert's output is multiplied by the gate value. Note that the computational cost is independent of the number of experts since a single expert is active for each position.}
  \label{fig:switch_transformer}
\end{figure}

\textbf{Mixture of Experts}~\citep{JacobsJordanNowlanEtAl91}: The core idea is to train multiple networks called experts, each of which specializes only in a subset of the data, and a manager or router, which forwards the input to the corresponding experts. A single network is used in practice, whose layers are composed of multiple subsets of parameters (experts), effectively resulting in a sparsely activated model as illustrated in Figure~\ref{fig:switch_transformer}. Increasing the number of experts keeps the computational cost constant since the model always selects the same number of experts for each input regardless of the number of experts. Therefore, the mixture of experts (MoE) approach allows for massive models and is particularly efficient for distributed systems in which experts are spread across devices. In that case, the number of experts, and therefore parameters, scales with the number of devices available. Despite these advantages, the mixture of experts has not yet been widely adopted as the method is complex to deploy in practice. It imposes a communication cost between the devices, a computation cost to select the experts for each input position, and makes training unstable. Recently, \citet{fedus2021switch} introduced the Switch Transformer based on a carefully crafted mixture of experts. Notably, given a fixed amount of computation per input position, the Switch Transformer reached the same quality threshold as a vanilla Transformer five times faster (wall-clock time) on average. Additionally, when trained further, the Switch Transformer outperformed the vanilla baseline. However, this approach assumes that multiple regimes with distinct input to output relations produce the data.

Difficult tasks often require large models to achieve the desired performance. However, such models require powerful and expensive accelerators. Both micro-batching and the mixture of experts offer an alternative to train such models on many relatively weak and inexpensive GPUs at the cost of complex implementation.

\textbf{Sample-Efficient Objective}~\citep{clark2020electra}: Large neural networks, especially Transformers, benefit from being pre-trained with an unsupervised objective before being fine-tuned on the task of interest, also called the downstream task. The core idea is to leverage large unlabelled datasets that are easy to automatically collect in order to learn the data underlying explanatory factors and ultimately improve the model performance. Concretely, pre-training initializes the network's weights in a ``good'' region of space.
As pre-training of large models is often more compute-intensive than fine-tuning, researchers regularly share pre-trained models to facilitate their adoption. Most notably, Hugging Face~\citep{wolf-etal-2020-transformers} is an open-source library that contains an extensive collection of pre-trained Transformers under a unified API. Nonetheless, researchers must sometimes pre-train models themselves due to the peculiar nature of the data or the problem at hand. In that case, a sample-efficient objective will reduce the computation required.

\begin{figure}[htb]
  \centering
  \includegraphics[scale=0.47]{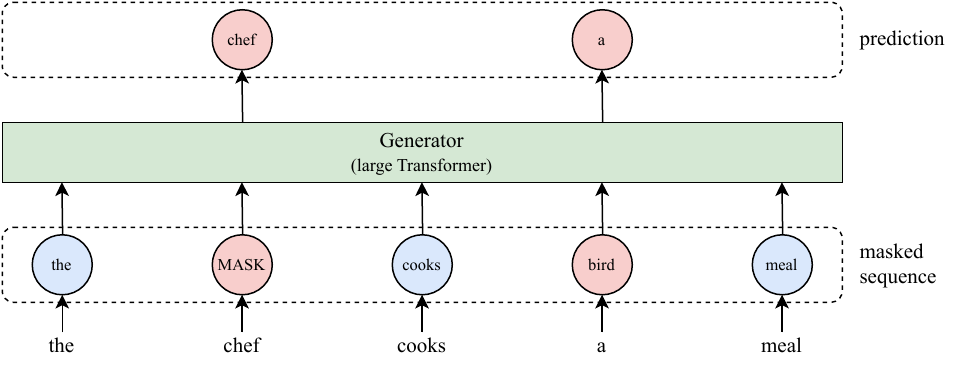}
  \Description[The masked language model objective.]{The masked language model objective.}
  \caption{The masked language model objective~\citep{devlin2019bert}. The masked words are depicted in red. The model makes a prediction only for the masked words; thus, MLM is computationally inefficient.}
  \label{fig:mlm}
\end{figure}

Recently, \citet{devlin2019bert} popularized the Cloze procedure~\citep{doi:10.1177/107769905303000401} for pre-training under the name of masked language model (MLM), which independently estimates the probability of masked words given the rest of the sequence. Practically, 15\% of the words are randomly selected, of which 80\% are masked, 10\% are replaced by a random word, and 10\% are left unchanged. This task is analogous to the reconstruction of corrupted input. Figure~\ref{fig:mlm} illustrates the masked language model objective.

\begin{figure}[htb]
  \centering
  \includegraphics[scale=0.47]{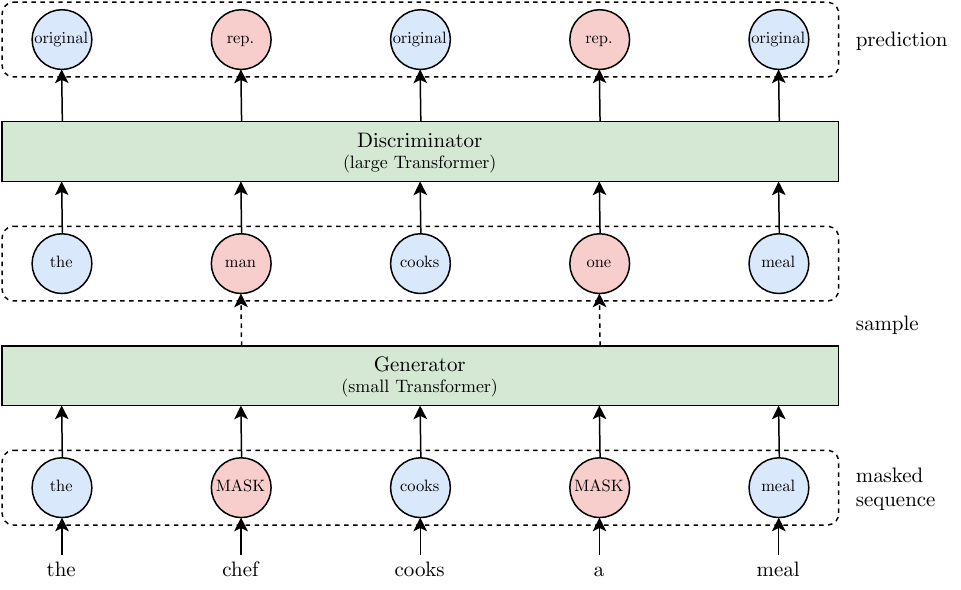}
  \Description[The replaced token detection objective.]{The replaced token detection objective.}
  \caption{The replaced token detection objective~\citep{clark2020electra}. A plausible alternative of each masked word is sampled from a small generator network. Then a discriminator predicts whether each word has been replaced.}
  \label{fig:replaced}
\end{figure}

\citet{clark2020electra} introduced the replaced token detection objective to speed up pre-training; a small network (generator) first generates a plausible alternative for each masked word, then the large model (discriminator) predicts whether each word has been replaced (see Figure~\ref{fig:replaced}). While the masked language model makes a prediction only for the masked works, the replaced token detection makes a prediction for every word. Therefore, the latter is more computationally efficient than the former; in other words, less pre-training computations are required to achieve the same performance on downstream tasks.  Additionally, the authors reported that the representations learned with their objective outperformed those learned with MLM given the same model size, data, and computation. Most notably, they were able to outperform GPT on the GLUE benchmark with 30$\times$ fewer computations.

\textbf{Parameter Initialization Strategies}: Optimizing deep networks is challenging in part because of the considerable influence of the initial point on the iterative process. Notably, the initial point determines whether the algorithms converge at all and, if it does converge, the speed at which it converges as well as the quality of the solution~\citep{Goodfellow-et-al-2016}. Transformers are notoriously difficult to train, typically requiring carefully tuned optimizers with adaptive learning rates, learning rate schedulers, and large batches. Even then, convergence is not guaranteed. Consequently, \citet{liu2020understanding} and \citet{huang2020improving} concurrently proposed initialization schemes for the Transformer that promise a smoother and faster optimization as well as better generalization performances.

\citet{liu2020understanding} identified an amplification effect that significantly influences training: each layer heavily depends on its residual branch\footnote{For a residual block $f(x) + x$, the residual branch refers to $f(x)$ and the skip connection, shortcut connection, or residual connection refers to $x$.}, making the optimization unstable as it amplifies small parameter perturbations. Ultimately, the amplification effect may produce a notable change in the Transformer's output. Nonetheless, the authors observed that heavy dependencies on the residual branches are necessary to unlock the Transformer's potential and achieve better results. In order to mitigate the amplification effect, \citet{liu2020understanding} introduced the Adaptive Model Initialization strategy, or Admin, that controls the dependency on the residual connections in the early stage of training with a new parameter $\boldsymbol{\omega}$. Formally, the $i$-th sub-layer output is given by
\begin{equation}
  \boldsymbol{X}_{i} = \mathrm{LayerNorm}(f_i(\boldsymbol{X}_{i-1}) + \boldsymbol{X}_{i-1} \odot \boldsymbol{\omega}_i),
\end{equation}
where $f_i(\boldsymbol{X})$, $\boldsymbol{X}_{i-1}$, and $\boldsymbol{X}_{i}$, denote the function, input, and output of the $i$-th sub-layer, respectively. Although this is equivalent to rescaling some model parameters, the authors observed that rescaling leads to unstable training in half-precision.

The proposed initialization strategy requires three steps. First, the model parameters are initialized with a standard method such as the Xavier initialization~\citep{pmlr-v9-glorot10a} and the Admin parameter $\boldsymbol{\omega}$ with ones. Then, one or a small number of mini-batches are forward propagated without updating the parameters and record the output variance of each residual branch $\mathrm{Var}[f_i(\boldsymbol{X}_{i-1})]$. Finally, the Admin parameter is initialized as $\boldsymbol{\omega}_i=\sqrt{\sum_{j<i}\mathrm{Var}[f_j(\boldsymbol{X}_{j-1})]}$. Once the model has been trained, $\boldsymbol{\omega}$ may be discarded.

The amplification effect is, however, not the only mechanism that makes Transformers notoriously difficult to train. \citet{huang2020improving} addressed two other issues: (i) Transformers are typically trained with optimizers that rely on adaptive learning rates as conventional SGD fails to train them effectively. However, adaptive learning rates have a problematically large variance in the early stages of optimization, resulting in convergence issues~\citep{Liu2020On}; and (ii) the magnitude of the error signal propagated through $\mathrm{LayerNorm}$ is inversely proportional to the magnitude of the input~\citep{pmlr-v119-xiong20b}. Specifically, the norm of the layer normalization gradient is proportional to:
\begin{equation}
  \left\lVert \frac{\partial\mathrm{LayerNorm}(\boldsymbol{x})}{\partial\boldsymbol{x}} \right\rVert = \mathcal{O}\left(\frac{\sqrt{d}}{\left\lVert \boldsymbol{x} \right\rVert}\right)
\end{equation}
Consequently, if the input norm $\left\lVert \boldsymbol{x} \right\rVert$ is larger than $\sqrt{d}$, backpropagating through layer normalization reduces the gradient magnitude for layers closer to the input. As a solution to both problems, \citet{huang2020improving} proposed an initialization strategy called T-Fixup that restricts the magnitude of the updates in the early stages of training, thus mitigating the vanishing gradient issue while eliminating the need for layer normalization and warmup.


While \citet{liu2020understanding} and \citet{huang2020improving} claim faster convergence, they did not report the improvement.

\textbf{Architecture Search}: One of the most challenging goals in deep learning is to automatically design networks. Indeed, the problem of finding architectures that achieve the best performance with the fewest operations and lowest memory footprint in a discrete search space is an NP-hard combinatorial optimization problem. Over the years, multiple approaches to Neural Architecture Search (NAS) have been proposed, including reinforcement learning~\citep{ZophL17}, evolutionary algorithms~\citep{RealAHL19}, and bilevel optimization~\citep{LiuSY19}. Notably, \citet{ZophVSL18} demonstrated that NAS is able to surpass human-designed architectures on ImageNet by 1.2\% top-1 accuracy while using 28\% fewer computations. Nonetheless, neural architecture search methods are computationally expensive as they usually require training each candidate model from scratch. As a solution, \citet{PhamGZLD18} proposed Efficient NAS (ENAS), which constrains all candidates to be subgraphs of a single computational graph, that is, to share parameters. Therefore, the ENAS's controller decides which operations are activated and relies on the models' ability to adapt, similarly to dropout~\citep{SrivastavaHKSS14}. Efficient NAS reduces the search computational budget by 1,000$\times$ over the original NAS~\citep{ZophL17}. Alternatively, \citet{LiuSY19} proposed the Differentiable Architecture Search (DARTS), which casts the NAS problem as a differentiable bilevel optimization problem. The first level consists of a continuous relaxation of the discrete search space using a $\mathrm{Softmax}$ function over a list of candidate operations, and the second level involves the model's weights. However, the bilevel formulation requires training the weights to convergence to evaluate the architecture gradient. To avoid this substantial cost, the authors made the approximation of taking a single gradient step of the weights for one gradient step of the architecture parameters. The authors obtained comparable performances to non-differentiable NAS methods on ImageNet in the mobile setting using only 4 GPU-days, compared to 3,150 for evolutionary algorithms~\citep{RealAHL19} and 2,000 for NAS~\citep{ZophVSL18}. Differentiable Architecture Search obtained comparable results to ENAS with a similar computational budget. We refer the reader to \citet{elsken2019} survey for further detail on architecture search methods.

Nevertheless, neural architecture search methods are challenging to apply on Transformers due to the memory requirements and training time. Therefore, recent works introduced methods better suited for the Transformer. \citet{SoLL19} modified the tournament selection evolutionary architecture search~\citep{RealAHL19} with Progressive Dynamic Hurdles (PDH), which dynamically allocates resources to more promising architectures according to their performances. With PDH, the authors optimized transformer architectures directly on the WMT'14 En-De task~\citep{bojar-EtAl:2014:W14-33} which requires 10 hours of computation on a Google TPU v2 for the base Transformer model. Training directly on this dataset is essential since the authors did not find a smaller surrogate dataset that transfers well, such as CIFAR-10 for ImageNet. The Evolved Transformer matched the vanilla Transformer's performance with only 78\% of its parameters. Recently, \citet{tsai2020} profiled the Transformer's components on a TPU v2 and observed that some mechanisms substantially impact inference time: attention queries, keys, and values dimensions, width and depth of feed-forward layers, number of attention heads, and layer normalization mean computation. By decomposing these components into building blocks and using binary variables, the authors perform a one-shot search for both the architecture and the parameters with a single loss. They optimized this loss with gradient descent on a continuous relaxation of the binary variables and used policy gradient algorithm. \citet{tsai2020} were able to make miniBERT 1.7$\times$ faster with a performance drop smaller than 0.3\%. Compared to the original BERT, this is 33 to 36$\times$ faster.

Neural architecture search is a promising tool to design lighter and faster Transformers automatically. Nonetheless, NAS imposes a high computational and memory cost, which may be avoided by carefully engineering the architecture instead. For instance, the Lite Transformer~\citep{wu2020lite} leverages the Long-Short Range Attention (LSRA), where a convolutional layer is applied in parallel to the self-attention in order to learn the local dependencies separately. The carefully handcrafted Lite Transformer outperforms the Evolved Transformer~\citep{SoLL19} for the mobile NLP setting while requiring about 14,000$\times$ less GPU time.

\textbf{Conditional Computing}~\citep{Bengio13}: Although large models are necessary for hard examples, smaller models are likely to perform as well, if not better, on simpler ones. For instance, many words such as ``car'' are easy to translate, while a few such as ``can'' require careful consideration of the context\footnote{Depending on the context, the word ``can'' has various meanings, including ``be able to'', ``may'', ``jail'', and ``metal container''. See \url{https://www.wordreference.com/definition/can}.}. As of this survey's writing, most architectures apply a fixed number of operations to all examples regardless of their difficulty. A more efficient approach would be to reduce the amount of computation for simple examples. As a solution, \citet{Bengio13} introduced conditional computing, which dynamically adapts the model's computational graph as a function of the input.

One way to implement conditional computing is with a mixture of experts, as introduced previously. In that case, only a subset of the parameters is used for a given input, making the computational graph sparse and the computation time almost constant with respect to the model size. Another approach consists of keeping the number of parameters constant and letting the model adjust its computation time separately for each input (according to the input's value). This approach is called Adaptive Computation Time (ACT)~\citep{Graves16} and uses a recurrent mechanism to transform the representations until a halting probability exceeds a given threshold. The model learns to control this probability to minimize both the prediction error and the number of iterations, called the \emph{ponder cost}, which prevents the model from using an infinite amount of computation before making a prediction. One shortcoming of the Adaptive Computation Time is its sensitivity to the ponder cost, which controls the trade-off between speed and accuracy.

\citet{dehghani2019universal} applied ACT to a Transformer with a recurrent mechanism for the architecture's depth. To implement this mechanism, the authors defined encoder and decoder blocks similar to the original Transformer, except that each block is recurrent, sending its output back as its input until the ponder cost becomes too high. Note that a fixed number of recurrent steps is equivalent to a Transformer with tied parameters across all layers. With this new architecture called Universal Transformer, the authors claimed that it is computationally universal (Turing-complete) given enough memory. This property may help Transformers generalize to sequences longer than the ones seen during training. The authors obtained state-of-the-art results on algorithmic and language understanding tasks. ACT and the Universal Transformer apply the same layers iteratively, which may not be sufficiently flexible. \citet{ElbayadGGA20} addressed this limitation with the  Depth-Adaptive Transformer (DAT), which applies different layers at every depth. The DAT matches the performance of a well-tuned Transformer baseline while reducing the computation by up to 76\%. However, the authors did not provide a comparison between the Universal Transformer and DAT.

In the same way that complex examples may require more computations, some may require access to a longer context. As a solution, \citet{sukhbaatar2019adaptive} dynamically adjusted the attention span, that is, the context length, by learning to mask the compatibility scores depending on the input. Their approach achieved state-of-the-art on \texttt{text8} and \texttt{enwik8}~\citep{enwik8} while requiring significantly fewer computations. Alternatively, \citet{li2020transformerbased} introduced the Decoder-end Adaptive Computation Steps (DACS), which monotonically computes halting probabilities along with the encoder states and stops the decoder computations in order to produce an output when the accumulation of probabilities exceeds a given threshold. In other words, each decoder step only looks at the necessary information as measured by the halting probabilities instead of looking at the entire input sequence.

\section{Specialized Approaches}
\label{sec:specialized}

Since the Transformer's quadratic complexity comes from the attention mechanism, most specialized methods rely on a fast and light approximation of the original full attention. As will be explained in greater detail in the rest of this section, the attention weight matrix is dominated by a few large values and is approximately low-rank. These observations justify two distinct lines of work: sparse attention and factorized attention. Alternatively, the complexity may be reduced without altering the original attention mechanism and thus the Transformer's capacity by directly modifying the network's architecture.
Let us first investigate the approaches that rely on sparse attention.

Note that some approaches only consider autoregressive tasks, such as the left-to-right language model, and in that case, the connectivity matrix is lower triangular as it is not permitted to attend to future positions. Whenever possible, such works have been extended to the more general case where attending to future positions is allowed in order to ease the comparison between the different approaches.

\subsection{Sparse Attention}
\label{ssec:sparse}

Due to the exponential nature of the $\mathrm{Softmax}$, only a few positions are strongly attended to. Consequently, a conceptually simple way of reducing the Transformer's complexity is to make the matrix $\boldsymbol{QK}^\top$ sparse\footnote{Since the matrix $\boldsymbol{QK}^\top$ is passed through a $\mathrm{Softmax}$ function, the masked values are set to minus infinity, effectively setting their contribution to $e^{-\infty}=0$.}, in other words, to only allow each position to attend to a subset of the positions. Let us investigate sparse patterns that are (i) fixed and random, (ii) learned and adaptive, and (iii) identified with clustering and locality sensitive hashing.

\begin{figure}[htb]
  \centering
  \includegraphics[scale=0.2]{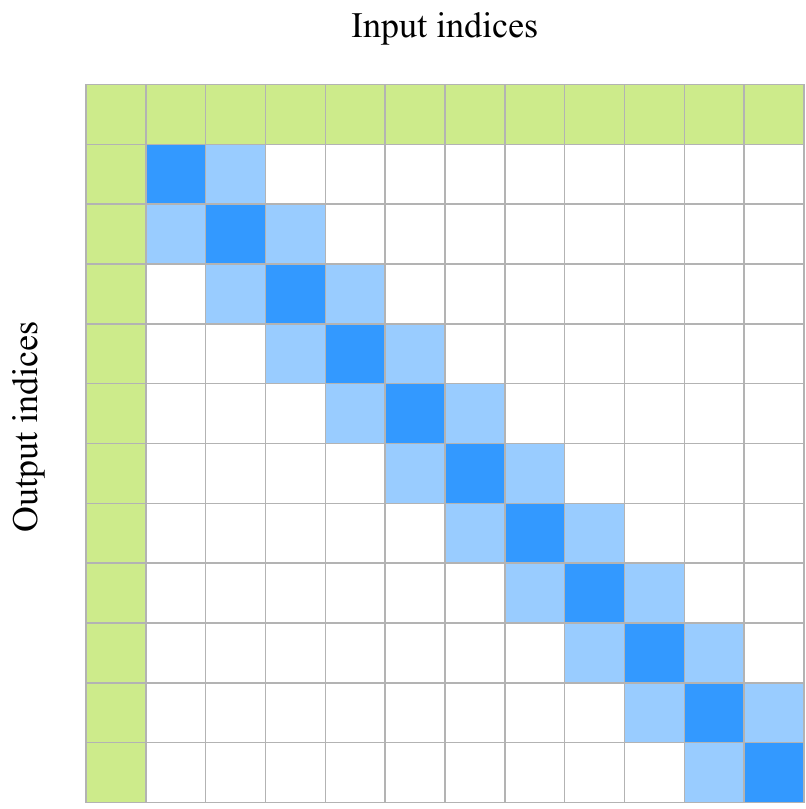}
  \Description[The connectivity matrices of the Star-Transformer.]{The connectivity matrices of the Star-Transformer.}
  \caption{The connectivity matrices of the Star-Transformer~\citep{guo2019startransformer}.}
  \label{fig:star}
\end{figure}

\textbf{Fixed and Random Sparse Patterns}~\citep{guo2019startransformer, DBLP:journals/corr/abs-1904-10509, wang2020transformer, li2020enhancing, qiu2020blockwise, 2020arXiv200405150B, NEURIPS2020_c8512d14}: One of the first models to consider fixed sparse patterns is the Star-Transformer introduced by \citet{guo2019startransformer}, which reduced the complexity from quadratic to linear by only allowing attention between adjacent positions. In order to preserve the Transformer’s ability to model long-term dependency, the authors relied on a single global token. Global tokens, also known as shared relay nodes, can attend to every position, and every position can attend to global tokens. Let us assume that the global token is located at position $0$. The $i$-th output position is allowed to attend to every input position if $i=0$, otherwise, it is allowed to attend to the $j$-th input positions for $j=0$ and if $i-1 \leq j \leq i+1$. Figure~\ref{fig:star} illustrates the Star-Transformer attention pattern.

\begin{figure}[htb]
  \centering
  \Description[The connectivity matrices of the Sparse Transformer.]{The connectivity matrices of the Sparse Transformer.}
  \includegraphics[scale=0.2]{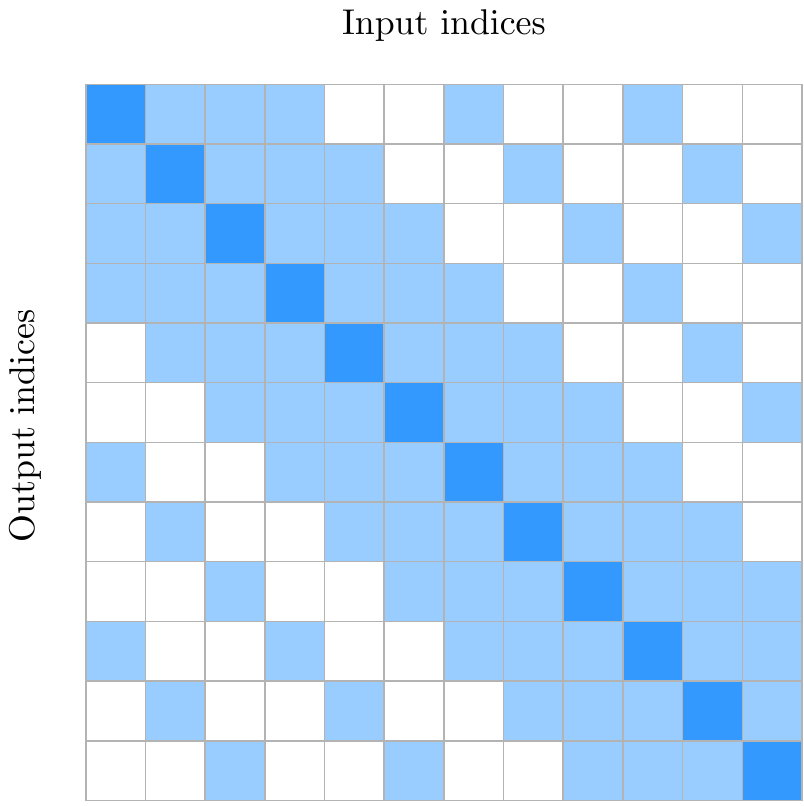}
  \includegraphics[scale=0.2]{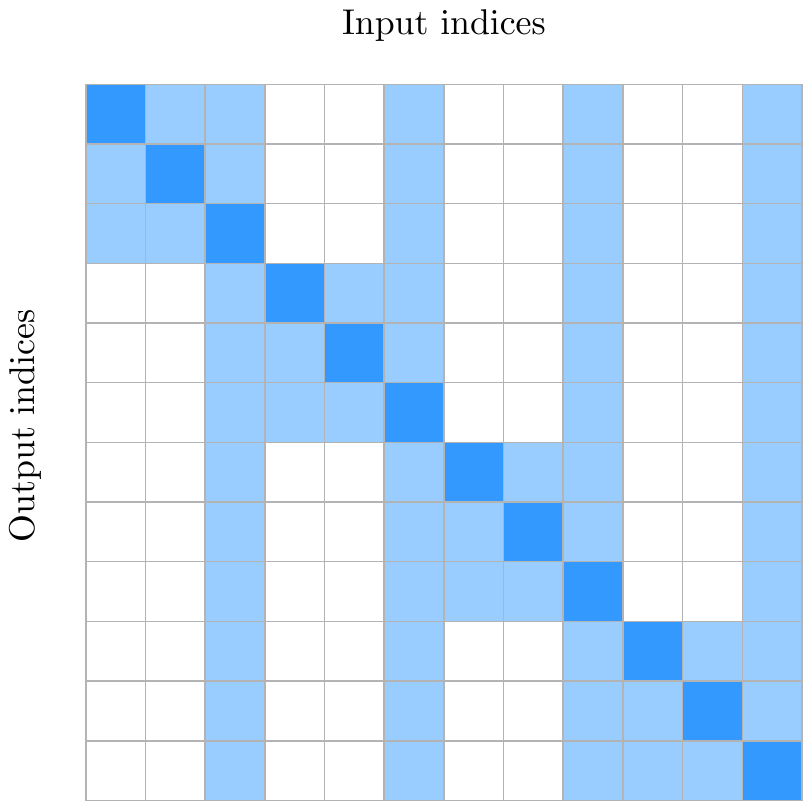}
  \caption{The connectivity matrices of the Sparse Transformer~\citet{DBLP:journals/corr/abs-1904-10509}. (Left) Strided attention with a stride of 3. (Right) Fixed attention with a stride of 3 and $c=1$.}
  \label{fig:sparse}
\end{figure}

Concurrently, \citet{DBLP:journals/corr/abs-1904-10509} introduced the Sparse Transformer which reduced the complexity to $\mathcal{O}(n\sqrt{n})$ with two different sparse attention patterns: strided and fixed. Strided attention allows the $i$-th output position to attend to the $j$-th input position if one of the two following conditions is satisfied: $(i + s) > j > (i - s)$ or $(i - j) \mathrm{\ mod\ }s = 0$, where the stride $s$ is chosen to be close to $\sqrt{n}$. Similarly, fixed attention allows $i$ to attend to $j$ if one of the two following conditions is satisfied: $\mathrm{floor}(j/s) = \mathrm{floor}(i/s)$ or $(j \mathrm{\ mod\ } s) \geq (s - c)$, where $c$ is an hyperparameter. Figure~\ref{fig:sparse} illustrates the strided and fixed attention patterns.

\begin{figure}[htb]
  \centering
  \Description[The connectivity matrices of the Cascade attention.]{The connectivity matrices of the Cascade attention.}
  \includegraphics[scale=0.2]{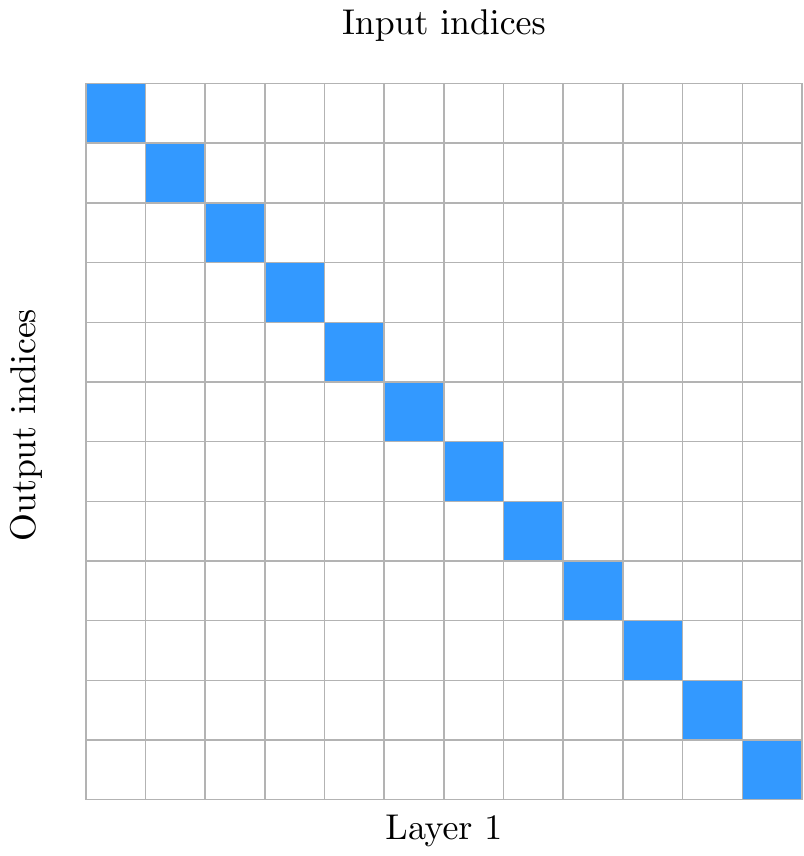}
  \includegraphics[scale=0.2]{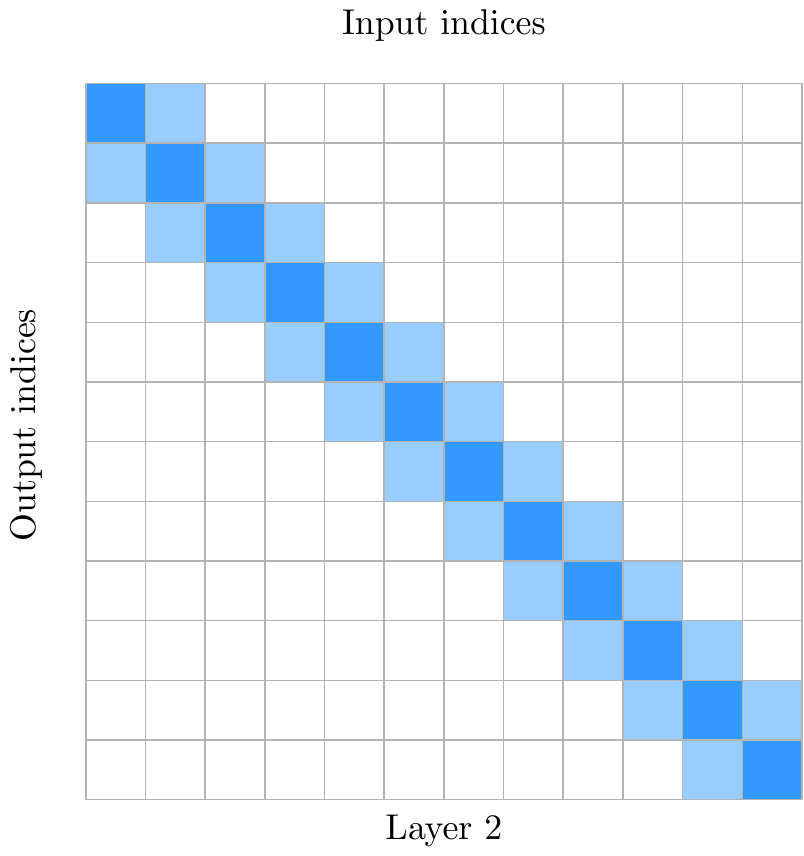}
  \includegraphics[scale=0.2]{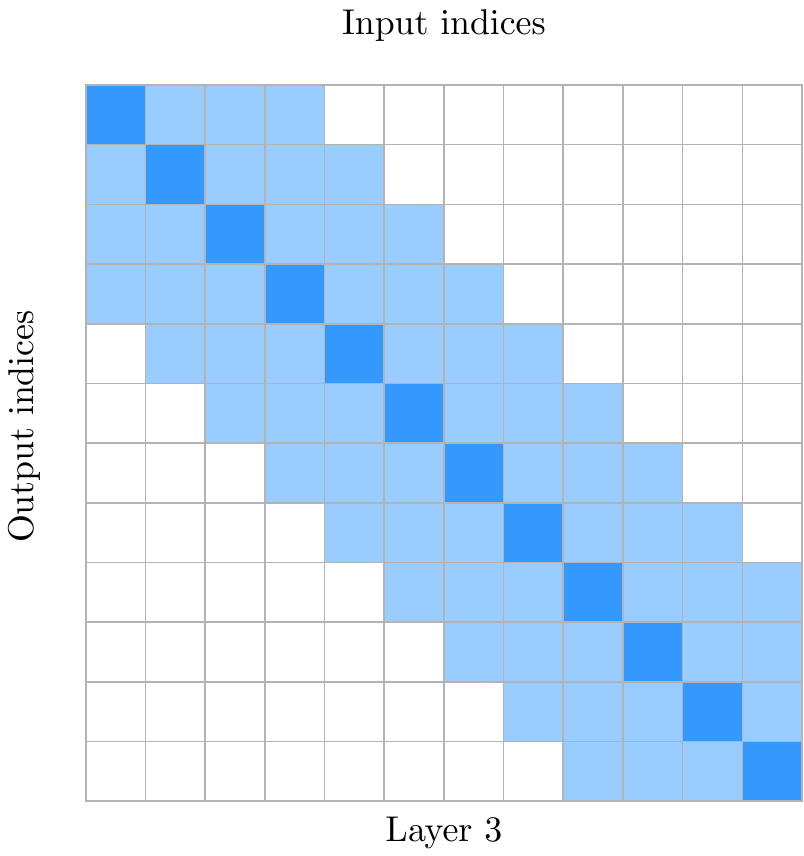}
  \includegraphics[scale=0.2]{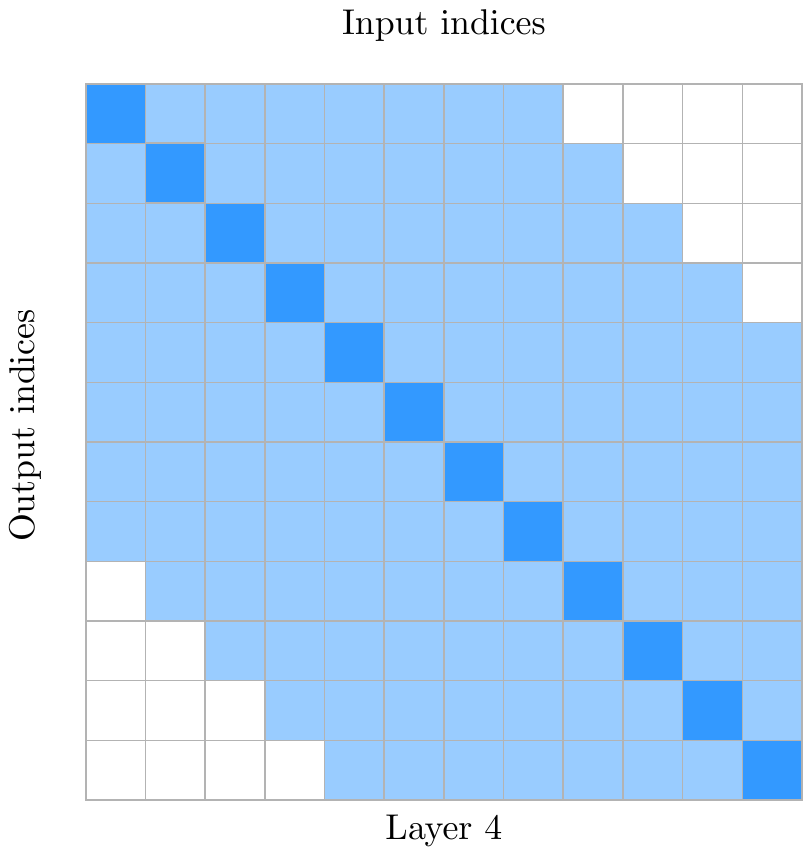}
  \caption{The connectivity matrices of the Cascade attention~\citep{wang2020transformer} for the first four layers with a base window $b=1$ and a cardinal number $m=2$. For instance, the window size of the third layer ($l=2$) is equal to $2\times b\times m^l - 1 = 7$.}
  \label{fig:sparse1}
\end{figure}

Alternatively, \citet{wang2020transformer} introduced the Cascade Transformer, which relies on sliding window attention whose size grows exponentially with the number of layers. More specifically, the number of cascade connections at the layer $l$ is equal to $2.b.m^l - 1$, where $b$ is the base window size and $m$ is the cardinal number; therefore reducing the complexity to $\mathcal{O}(n.b.m^l)$. Cascade attention is well suited for shallow networks, but its complexity tends to that of the full attention in deep networks as depicted by the connectivity matrices in Figure~\ref{fig:sparse1}.

\begin{figure}[htb]
  \centering
  \Description[The connectivity matrix of the LogSparse attention.]{The connectivity matrix of the LogSparse attention.}
  \includegraphics[scale=0.2]{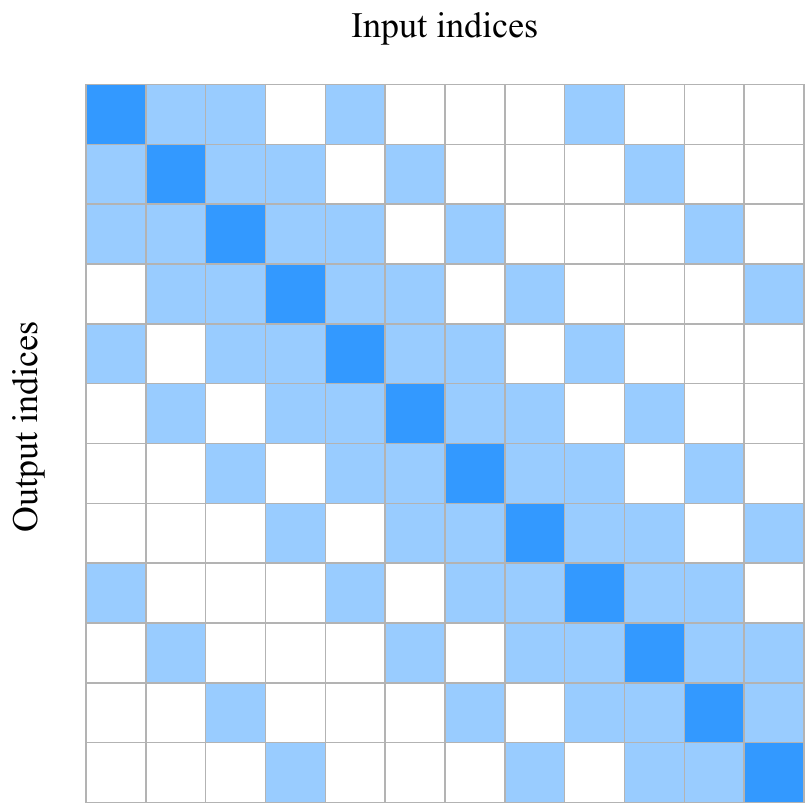}
  \caption{The connectivity matrix of the LogSparse attention~\citet{li2020enhancing}.}
  \label{fig:logsparse}
\end{figure}

\citet{li2020enhancing} introduced the LogSparse-Transformer for forecasting fine-grained time series with strong long-term dependencies. The LogSparse-Transformer relies on the eponym attention that allows the $i$-th output to attend to the $j$-th inputs for $j\in \{-2^{\lfloor\log_2 i\rfloor}, i-2^{\lfloor\log_2 i\rfloor-1}, \dots,i-2^1, i-2^0, i, i+2^0, i+2^1, \dots, i+2^{\lfloor\log_2 (n-i)\rfloor-1}, i+2^{\lfloor\log_2 (n-i)\rfloor}\}$ where $\lfloor.\rfloor$ denotes the floor operation and $N$ denotes the sequence length. Figure \ref{fig:logsparse} illustrates the connectivity matrix of the LogSparse attention. Since only $O(\log n)$ positions are attended to by each of the $n$ positions, the complexity of the LogSparse attention is $O(n\log n)$. Additionally, the authors proposed two alternatives: (1) to allow the $i$-th output to attend to the first $k$ input positions, after which the LogSparse attention is resumed, and (2) to divide the input sequence into subsequences, and to apply the LogSparse attention on each of them.

\begin{figure}[htb]
  \centering
  \Description[The connectivity matrices of the block-wise attention.]{The connectivity matrices of the block-wise attention.}
  \includegraphics[scale=0.2]{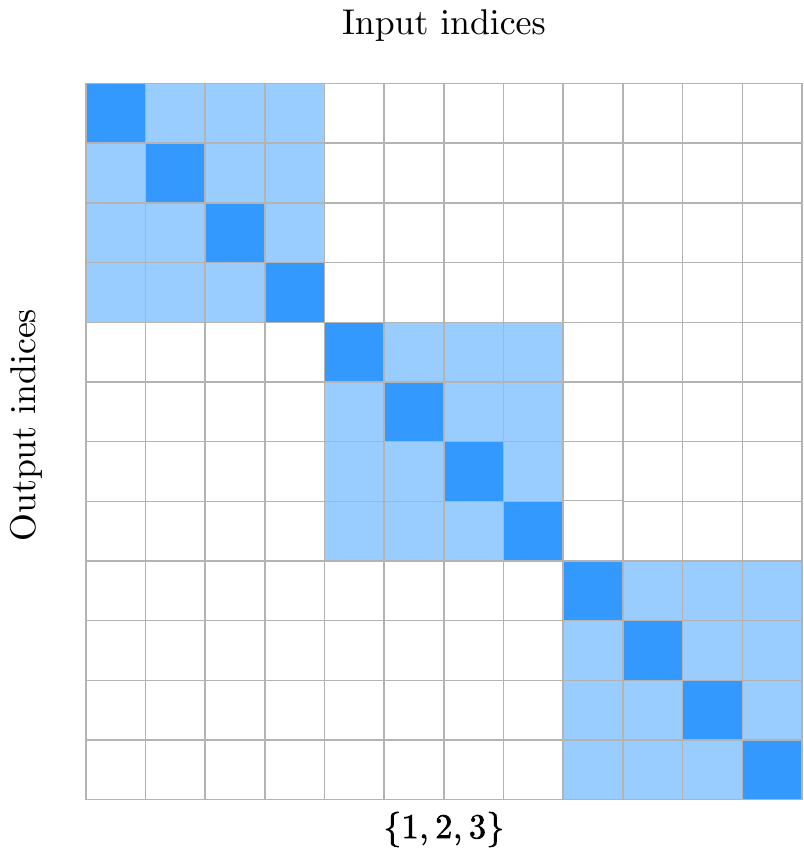}
  \includegraphics[scale=0.2]{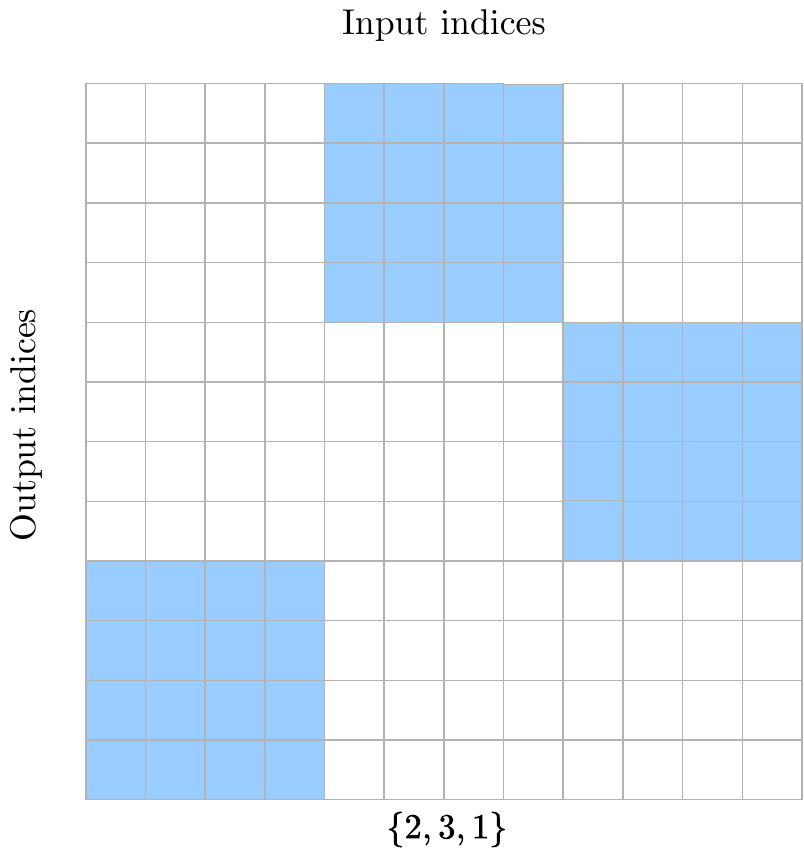}
  \includegraphics[scale=0.2]{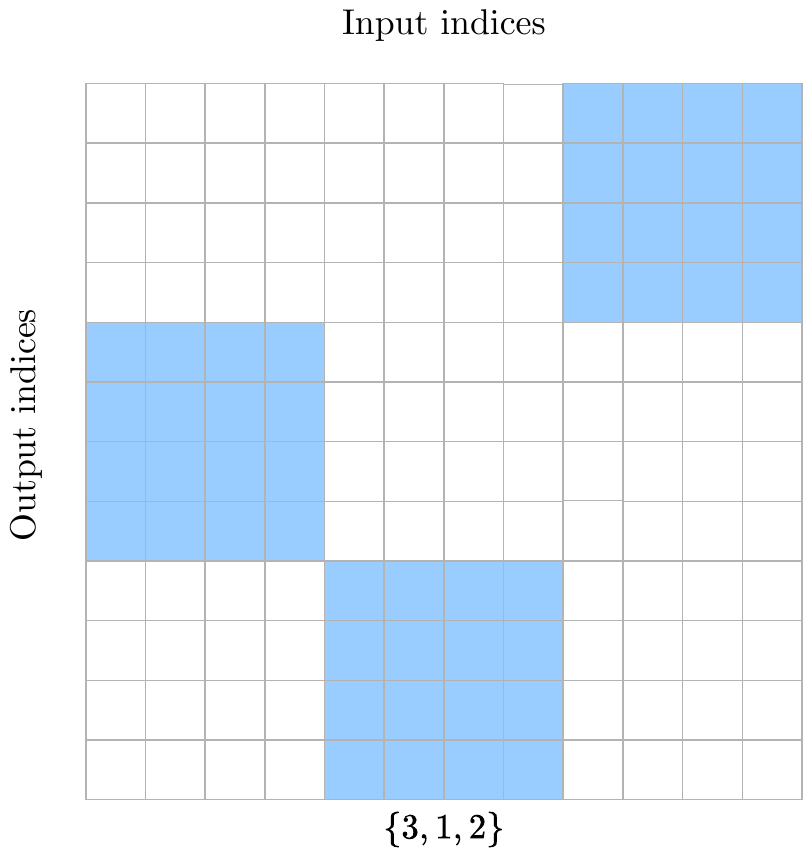}
  \caption{The connectivity matrices of the block-wise attention~\citep{qiu2020blockwise} for $n_b=3$ blocks. The corresponding permutations are written below the connectivity matrices.}
  \label{fig:sparse3}
\end{figure}

\citet{qiu2020blockwise} introduced BlockBERT, which relies on the block-wise attention: the input sequence is split into $n_b$ non-overlapping blocks, and positions in block $i$ are only allowed to attend to positions in block $\pi(i)$, where $\pi$ denotes a permutation. The author chose to generate the permutations by simply shifting the positions. For instance, the possible permutations of $\{1, 2, 3\}$ are $\{1, 2, 3\}$, $\{3, 1, 2\}$, and $\{2, 3, 1\}$. The permutation $\{2, 3, 1\}$ means that the first block attends to the second block, the second block attends to the third block, and the third block attends to the first block. In the multi-head setting, a different permutation\footnote{Note that if the number of heads is greater than the number of permutations, multiple heads must be assigned the same permutation.} is assigned to each head. More formally, the output position $i$ is only allowed to attend to input $j$ if the following condition is satisfied:
\begin{equation}
  \pi\left(\left\lfloor\frac{(i-1)n_b}{n} + 1\right\rfloor\right) = \left\lfloor\frac{(j-1)n_b}{n} + 1\right\rfloor
\end{equation}
\noindent Figure~\ref{fig:sparse3} illustrates the connectivity matrix of the block-wise attention where a sequence of length $n=12$ is split into $n_b=3$ blocks. Although the block-wise attention reduces the memory and computational cost by a factor $n_b$, the complexity remains quadratic with respect to the sequence length.

\begin{figure}[htb]
  \centering
  \Description[The connectivity matrices of the Longformer and BigBird.]{The connectivity matrices of the Longformer and BigBird.}
  \includegraphics[scale=0.2]{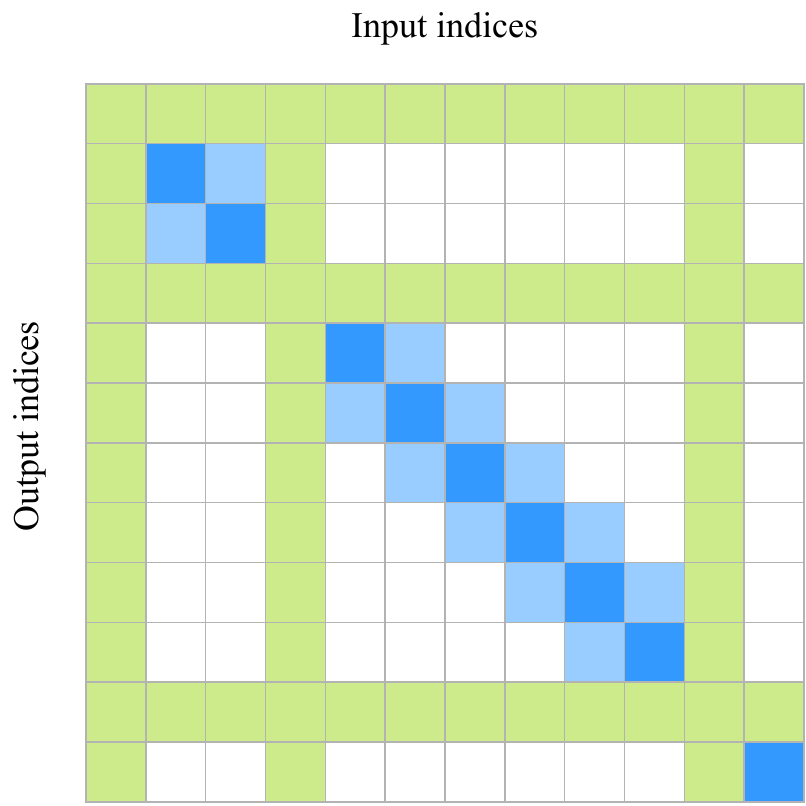}
  \includegraphics[scale=0.2]{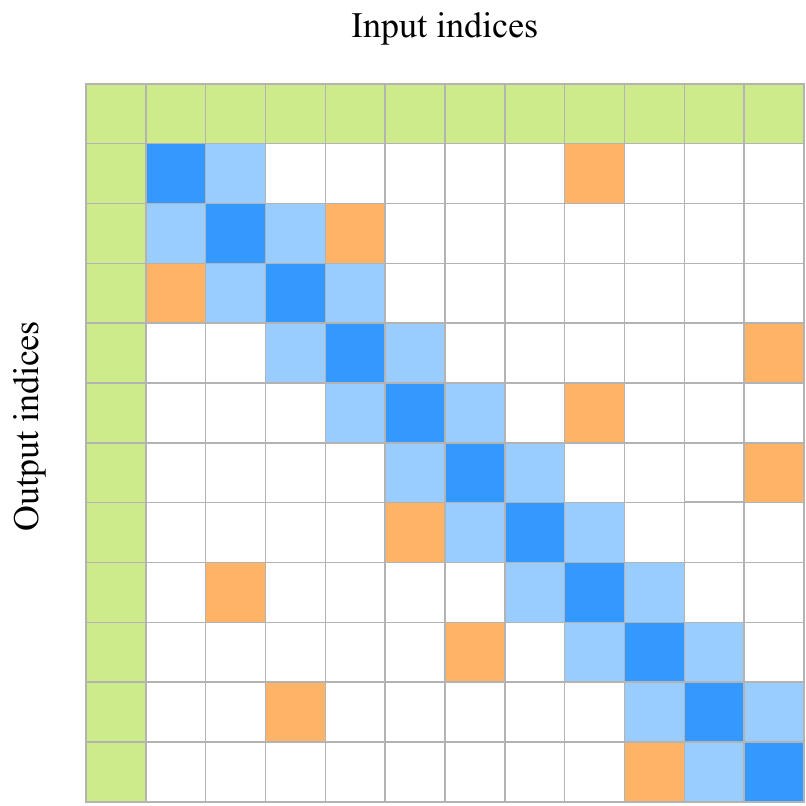}
  \caption{The connectivity matrices of two sparse attention schemes. (Left) Longformer~\citep{2020arXiv200405150B}. (Right) BigBird~\citep{NEURIPS2020_c8512d14}. The attention is the combination of sliding window attention (blue), global attention (green), and random attention (orange).}
  \label{fig:sparse2}
\end{figure}

\citet{2020arXiv200405150B} introduced the Longformer which further reduces the complexity to $\mathcal{O}(n)$ using a combination of sliding window and global attentions (see Figure~\ref{fig:sparse2}). The assumption behind the sliding window attention is that the most useful information is located in each position's neighbourhood. The sliding window attention is limited in that it requires $\mathcal{O}(\sqrt{n})$ layers to model long-range dependencies. Thus, a few preselected tokens have a global attention: they can attend to every position and be attended by every position. Consequently, the maximum path length between any two positions is equal to~2. \citet{NEURIPS2020_c8512d14} introduced BigBird, which also achieves a linear complexity using a combination of random, sliding window, and global attentions (see Figure~\ref{fig:sparse2}). BigBird has two configurations that the authors referred to as internal transformer construction (ITC) and extended transformer construction (ETC). Similarly to the Longformer, the former uses existing positions for global attention, while the latter uses additional tokens, increasing the model's capacity and performance. Interestingly, the extra location of ETC may be seen as a form of memory. The authors proved that their sparse factorization preserves the theoretical properties of Transformers with the full attention: the model is both a universal approximator of sequence functions and Turing complete. However, BigBird without random attention outperformed BigBird with it in most of their experiments.

\textbf{Learned and Adaptive Sparse Patterns}~\citep{tay2020sparse, shi2021sparsebert, correia-etal-2019-adaptively}: Fixed and random patterns are handcrafted and may not be suitable for the data and task at hand. One may instead learn the relevant patterns and adapt them based on the content.

\begin{figure}[htb]
  \centering
  \Description[The connectivity matrix of the sparse Sinkorn attention.]{The connectivity matrix of the sparse Sinkorn attention.}
  \includegraphics[scale=0.2]{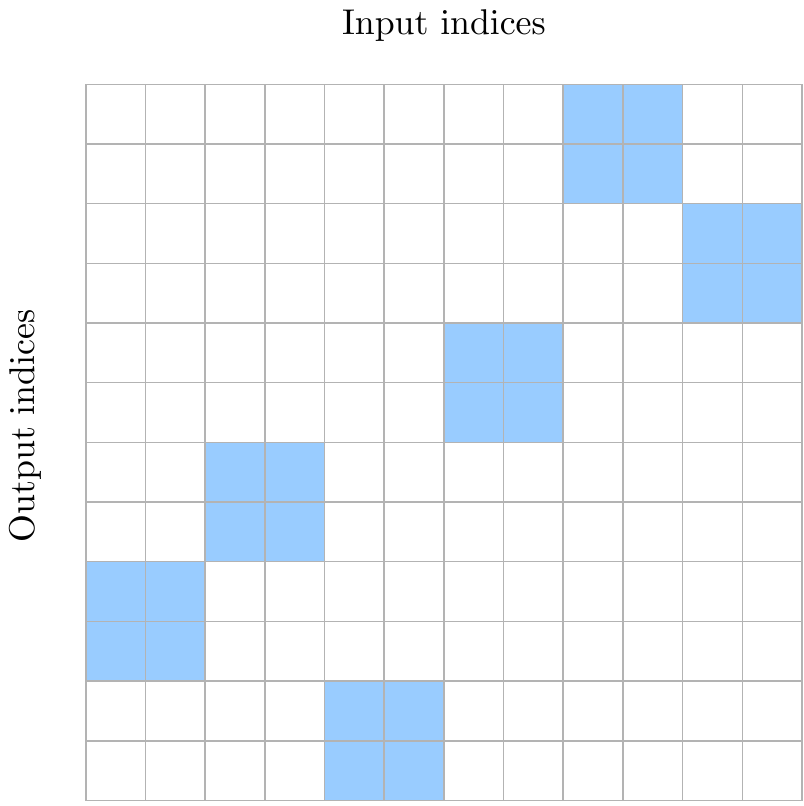}
  \caption{The connectivity matrix of the sparse Sinkorn attention~\citep{tay2020sparse}.}
  \label{fig:sparse4}
\end{figure}

In order to increase the flexibility of the block-wise attention, \citet{tay2020sparse} introduced the sparse Sinkhorn attention, which is equivalent to the block-wise attention whose keys have been sorted in a block-wise fashion. In other words, the permutations are learned. More specifically, the sparse Sinkhorn attention transforms the input sequence $\boldsymbol{X} \in \mathbb{R}^{n \times d}$ into $\boldsymbol{X}' \in \mathbb{R}^{n_b \times d}$ where $n_b$ is the number of blocks, and where $\boldsymbol{X}'_i$ is equal to the sum of the input in that block. A simple feed-forward network then learns a mapping $\boldsymbol{R}_i \in \mathbb{R}^{n_b}$ from the $i$-th block $\boldsymbol{X}'_i$ to all blocks. In order to obtain a sorting matrix from $\boldsymbol{R} \in \mathbb{R}^{n_b \times n_b}$, that is, a matrix comprising only $0$s and $1$s, and whose rows and column sum to one, the rows and columns are iteratively normalized. The sorting matrix is then used to permute the keys, effectively learning which block to attend (see Figure~\ref{fig:sparse4}). The sparse Sinkhorn attention reduces the complexity to $\mathcal{O}(n_b^2)$. Nonetheless, since the block size is constant in the original paper, the complexity remains quadratic with respect to the sequence length. Additionally, the authors proposed a truncated version of the sparse Sinkhorn attention, which selects a few keys after sorting them, further reducing the complexity to $\mathcal{O}(n)$.

Recently, \citet{shi2021sparsebert} put under the microscope the attention patterns learned by BERT~\citep{devlin2019bert} and observed that the diagonal elements are less important compared to other positions, that is, they contribute the least to the output, while neighbourhood positions and special tokens are prominent. To confirm their observations, they dropped the diagonal element in BERT's attention such that each position is not allowed to attend to itself and noted that the performance remains comparable to the original model. Additionally, they observed that models for different tasks have various degrees of redundancy and hence can achieve various sparsity levels before significantly dropping performance. Consequently, \citet{shi2021sparsebert} proposed to learn sparsity patterns for each task in an end-to-end fashion with the Differentiable Attention Mask (DAM) algorithm. Let us denote the attention score between the $i$-th output position (query) and $j$-th input position (key) as $\alpha_{i, j}$. They proposed to compute the attention mask $M_{i, j}$ as the $\mathrm{Gumbel}\mhyphen\mathrm{Sigmoid}$~\citep{DBLP:conf/iclr/MaddisonMT17} of the attention score $\alpha_{i, j}$:
\begin{equation}
  M_{i, j} = \mathrm{Gumbel}\mhyphen\mathrm{Sigmoid}(\alpha_{i, j})= \mathrm{Sigmoid}\left(\frac{\alpha_{i, j} + G_1 - G_2}{\tau}\right)
\end{equation}
where $G_1$, $G_2$ are independent Gumbel noises $G_k = -\log(-\log(U_k))$ generated from a uniform distribution $U_k \sim \mathcal{U}(0,1)$, and where $\tau$ is a temperature hyperparameter. Note that the $\mathrm{Gumbel}\mhyphen\mathrm{Sigmoid}$ becomes binary as $\tau$ approaches $0$. A penalty term $\lambda \lVert M\lVert _1$ is added to the loss to control the trade-off between performance and sparsity. The resulting model called SparseBERT achieved 91.2\% sparsity while maintaining an average score of 80.9\% on GLUE, i.e., only 3\% lower than the full BERT. Such an approach deviates from previous sparse attention whose patterns have been manually handcrafted. To avoid learning completely unstructured sparsity patterns, the authors proposed to enforce the first and last row/column of the attention mask to be active and all positions on each line parallel to the diagonal to share their mask parameters.

As mentioned above, due to the exponential nature of the $\mathrm{Softmax}$, most positions are lightly attended to. In other words, most attention weights are small but non-zero. Instead, \citet{correia-etal-2019-adaptively} introduced the Adaptively Sparse Transformer that replaces the $\mathrm{Softmax}$ by the $\alpha\mhyphen\mathrm{entmax}$ function, a differentiable generalization of the $\mathrm{Softmax}$ that pushes small weights to be exactly zero. Formally, the $\alpha\mhyphen\mathrm{entmax}$ function is defined as:
\begin{equation}
  \alpha\mhyphen\mathrm{entmax}(\boldsymbol{z}) = \underset{\boldsymbol{p} \in \Delta^d}{\mathrm{argmax\ }}\boldsymbol{p}^\top \boldsymbol{z} + \boldsymbol{H}^T_{\alpha}(\boldsymbol{p}),
  \label{eq:entmax}
\end{equation}
where $\Delta^d = \{\boldsymbol{p}\in \mathbb{R}^d : \sum_i p_i = 1\}$ and, for $\alpha \geq 1$, $\boldsymbol{H}^T_{\alpha}$ is the Tsallis continuous family of entropies:
\begin{equation}
  \boldsymbol{H}^T_{\alpha}(\boldsymbol{p}) = \begin{cases}
    \frac{1}{\alpha(\alpha - 1)}\sum_j (p_j - p_j^\alpha), & \alpha \ne 1 \\
    -\sum_j p_j \log p_j,                                  & \alpha = 1.
  \end{cases}
\end{equation}
The authors showed that the solution to the equation \ref{eq:entmax} is
\begin{equation}
  \alpha\mhyphen\mathrm{entmax}(\boldsymbol{z}) = \left[(\alpha-1)\boldsymbol{z} - \lambda \boldsymbol{1} \right]_{+}^{\frac{1}{\alpha - 1}},
\end{equation}
where $[]_{+}$ denotes the $\mathrm{ReLU}$ function, $\boldsymbol{1}$ denotes the vector of ones, and $\lambda$ is the Lagrange multiplier corresponding to the $\sum_i p_i=1$ constraint.

Interestingly, when $\alpha=1$, the $\alpha\mhyphen\mathrm{entmax}$ is equivalent to the $\mathrm{Softmax}$, and the attention is dense, and when $\alpha>1$, the output is permitted to be sparse. In their experiments, a scalar parameter $a_{i, j}$ is learned for the $j$-th attention head of the $i$-th layer, and $\alpha_{i, j}$ is computed as:
\begin{equation}
  \alpha_{i, j} = 1 + \mathrm{sigmoid}(a_{i, j}) \in\ ]1, 2[
\end{equation}
Nonetheless, the Adaptively Sparse Transformer computes the attention score for each pair of queries and keys. Consequently, the sparsity cannot be leveraged to improve the memory and computation, resulting in a model that is 25\% slower than the original Transformer in terms of tokens per second.

As of this survey's writing, unstructured sparse attention (whether fixed, random or learned) does not benefit from efficient implementations and therefore cannot result in memory and computational improvements. Nonetheless, there are exciting researches in that direction, as noted by \citet{DBLP:journals/corr/abs-2009-06489}. In contrast, some structured sparsity patterns benefit from efficient implementations. Recently, NVIDIA introduced its Ampere architecture which efficiently compresses 2:4 structured sparsity on rows, that is, two non-zero values in every four entries.

\textbf{Clustering and Locality-Sensitive Hashing}~\citep{Kitaev2020Reformer, roy2020efficient}: The $\mathrm{Softmax}$ function is dominated by the largest values, that is, by the keys and queries that have the largest dot product. Therefore, the attention may be approximated by only comparing the most similar keys and queries. Although this approach is a form of adaptive sparsity as the patterns depend on the data, they are presented separately due to their conceptual difference.

\begin{figure}[htb]
  \centering
  \Description[The connectivity matrix of the Reformer.]{The connectivity matrix of the Reformer.}
  \includegraphics[scale=0.2]{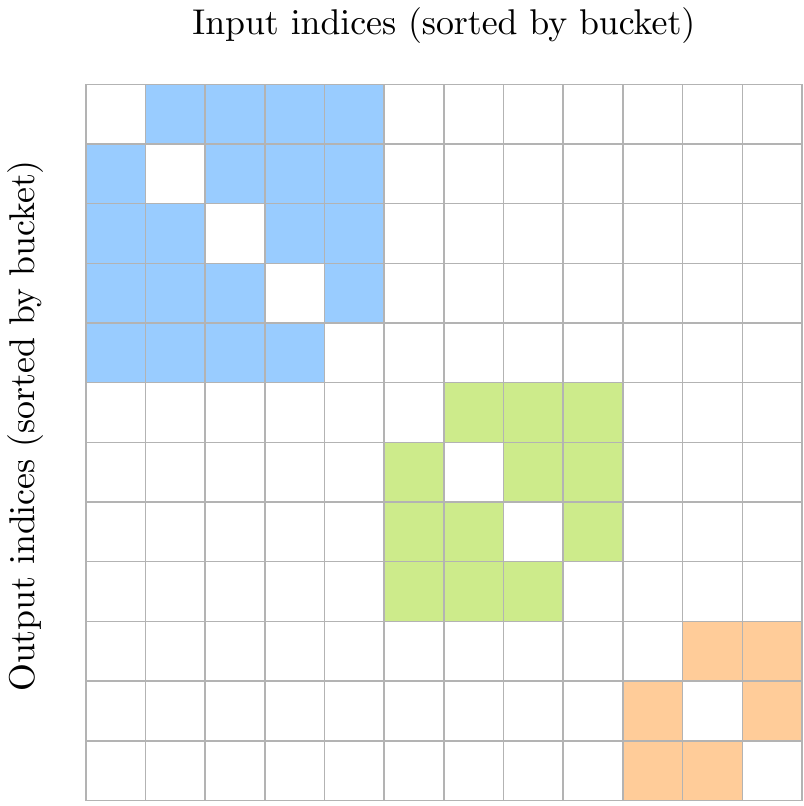}
  \caption{The connectivity matrix of the Reformer~\citep{Kitaev2020Reformer}. Queries and keys are bucketed using LSH then sorted by their bucket. Therefore, the $i$-th row of the connectivity matrix may not correspond to the $i$-th position in the input sequence. Units can only attend other units in the same bucket, but not themselves because queries and keys are equal. The colour represents buckets.}
  \label{fig:reformer}
\end{figure}

\citet{Kitaev2020Reformer} introduced the Reformer, which selects the set of keys that the query can attend to by grouping them with an angular multi-round locality-sensitive hashing (LSH). Such hashing scheme has a high probability of assigning the same value to similar vectors. Formally, queries and keys are shared ($Q=K$) and bucketed using $b$ hash values obtained as follows:
\begin{gather}
  \boldsymbol{p} = [\boldsymbol{x}^\top \boldsymbol{R};-\boldsymbol{x}^\top \boldsymbol{R}]\\
  h(\boldsymbol{x})=\underset{i}{\mathrm{argmax}}(p_i)
\end{gather}
\noindent where $;$ denotes the concatenation operation, and where $\boldsymbol{x} \in \mathbb{R}^{d}$ is a query/key and $\boldsymbol{R} \in \mathbb{R}^{d\times b/2} $ is a random rotation matrix. Output positions are only allowed to attend to input positions that are in the same bucket (see Figure~\ref{fig:reformer}). They are, however, not allowed to attend to themselves because the dot product of a vector with himself will almost always be greater than the dot product with other positions.

The authors chose a constant bucket size $l_B$, resulting in a number of buckets $n_B=n/l_B$. The attention complexity is $\mathcal{O}(n_B \times l_B^2)$ which simplifies as $\mathcal{O}(n)$. This does not take into account the computation of the hash values for each position. As only $\log n_B$ bits are required to encode $n_B$ buckets, the complexity of computing hash values is given by $\mathcal{O}(n \log n_B)$, which simplifies as $\mathcal{O}(n \log n)$. Consequently, the complexity of the Reformer's attention is  $\mathcal{O}(n \log n)$.

The Maximum Inner Product Search (MIPS) problem is the task of searching for the vector $K_j$ in $K=\{K_1, K_2, \cdots, K_n\}$ that maximizes the dot product with a given vector $Q_i$. Note that the MIPS problem is particularly useful for the attention mechanism as $Q_i^\top K_j$ is directly proportional to the contribution of the $j$-th value for the $i$-th attention's output. There are multiple approaches to approximately solve this problem, including tree-based and LSH-based. When the norm of every $K_j$ is constant, the problem is equivalent to the Nearest Neighbour Search (NNS). Motivated by this observation and to avoid the computational cost of learning sparsity patterns, \citet{roy2020efficient} proposed the Routing Transformer that relies on an online mini-batch version of $k$-means and a set of centroids learned along the rest of the parameters. Like the Reformer, queries can only attend to keys from the same cluster, inducing an adaptive or content-based sparsity pattern.

\subsection{Factorized Attention}
\label{ssec:factorized}

\citet{2020arXiv200604768W} demonstrated that the attention matrix $\mathrm{Softmax}\left(\boldsymbol{QK}^\top/\sqrt{d}\right)$ is approximately low rank. Consequently, another approach to reduce the Transformer's complexity is to approximate the attention by factorizing it into the product of two matrices with lower dimensions.

\textbf{Low-Rank Factorization}~\citep{2020arXiv200604768W, tay2020synthesizer, xiong2021nystromformer}: \citet{2020arXiv200604768W} introduced the Linformer, a linear complexity model that approximates the attention with a low-rank factorization by first projecting each key to a lower dimension before performing the dot product, thereby saving time and memory. Formally, the low-rank attention is given by:
\begin{equation}
  \mathrm{Attention}(\boldsymbol{X}) =\underbrace{\mathrm{Softmax}\bigg(\frac{\boldsymbol{QK}^\top}{\sqrt{d}}\bigg)}_{n \times n}\underbrace{\vphantom{\bigg(} \boldsymbol{V}}_{n \times d} \approx \underbrace{\mathrm{Softmax}\bigg(\frac{\boldsymbol{Q}(\boldsymbol{EK})^\top}{\sqrt{d}}\bigg)}_{n \times k}\underbrace{\vphantom{\bigg(} \boldsymbol{FV}}_{k \times d}
\end{equation}
\noindent where $\boldsymbol{E}, \boldsymbol{F} \in \mathbb{R}^{k \times n}$, with $k \ll n$, are two linear projection matrices learned during training. The authors showed that $\boldsymbol{E}$ and $\boldsymbol{F}$ could be shared across heads and layers with virtually no performance penalty.

\citet{tay2020synthesizer} introduced a family of models called Synthesizers that learn the compatibility scores without computing the pairwise dot products between the queries and keys. For instance, the Dense Synthesizer learns the compatibility scores with a simple position-wise feed-forward network that projects each of the $n$ rows of $\boldsymbol{X}$ from $\mathbb{R}^{1 \times d}$ to $\mathbb{R}^{1 \times n}$:
\begin{equation}
  \mathrm{F}(\boldsymbol{X}_i)=\max(0, \boldsymbol{X}_i\boldsymbol{W}_1+\boldsymbol{b}_1)\boldsymbol{W}_2 + \boldsymbol{b}_2
\end{equation}
\noindent where $\boldsymbol{W}_1 \in \mathbb{R}^{d \times d}$ and $\boldsymbol{W}_2 \in \mathbb{R}^{d \times n}$. Finally, the attention is given by:
\begin{equation}
  \mathrm{Attention}(\boldsymbol{X}) = \mathrm{Softmax}(F(\boldsymbol{X}))G(\boldsymbol{X})
\end{equation}
\noindent where $G(\cdot): \mathbb{R}^{n \times d} \rightarrow \mathbb{R}^{n \times d}$ is a projection of the input akin to the values. In order to improve the efficiency, the authors proposed the Factorized Dense Synthesizer which first project the input $\boldsymbol{X}$ with two feed-forward networks:
\begin{equation}
  \boldsymbol{A} = F_A(\boldsymbol{X}) \in \mathbb{R}^{n \times a} \quad \mathrm{and} \quad \boldsymbol{B} = F_B(\boldsymbol{X}) \in \mathbb{R}^{n \times b},
\end{equation}
\noindent such that $a \times b = n$. Then, two tiling functions $H_A(\cdot):\mathbb{R}^{n\times a} \rightarrow \mathbb{R}^{n\times (a.b)}$ and $H_B(\cdot): \mathbb{R}^{n\times b} \rightarrow \mathbb{R}^{n\times (b.a)}$ are applied to $\boldsymbol{A}$ and $\boldsymbol{B}$, respectively. Note that a tiling function simply repeats a vector multiple times. Finally, the attention of the Factorized Dense Synthesizer is given by:
\begin{align}
   & \mathrm{Attention}(\boldsymbol{X}) = \mathrm{Softmax}(H_A(\boldsymbol{A})H_B(\boldsymbol{B})^\top)G(\boldsymbol{X})
\end{align}
Additionally, the authors proposed a baseline called the Factorized Random Synthesizer, whose compatibility scores are independent of the input. Formally, the Factorized Random Synthesizer's attention is given by:
\begin{equation}
  \mathrm{Attention}(\boldsymbol{X}) = \mathrm{Softmax}(\boldsymbol{R}_1 \boldsymbol{R}_2^\top)G(\boldsymbol{X})
\end{equation}
\noindent where $\boldsymbol{R}_1, \boldsymbol{R}_2 \in \mathbb{R}^{n \times k}$ are two low-rank matrices learned during training. Although the Synthesizers eliminate the need to compute the pairwise dot products, which speed up the model in practice, the complexity remains quadratic with respect to the sequence length.

The Nyströmformer~\citep{xiong2021nystromformer} relies on the Nyström method to generate a low-rank approximation of the $\mathrm{Softmax}$ matrix. However, applying the Nyström method directly to the $\mathrm{Softmax}$ would require to compute the $\boldsymbol{QK}^\top$ product, which requires $\mathcal{O}(n^2)$ computations and memory. As a solution, the Nyströmformer creates two subsets $\tilde{\boldsymbol{K}}$ and $\tilde{\boldsymbol{Q}}$ of columns, called landmarks, from $\boldsymbol{K}$ and $\boldsymbol{Q}$, respectively. The authors applied the segment-means approach, which computes the landmarks as the averages over predefined spans of keys and queries. Let $\boldsymbol{S}_{AB}$ denotes $\mathrm{Softmax}(\boldsymbol{A}\boldsymbol{B}^\top/\sqrt{d})$ for any matrix $\boldsymbol{A}$ and $\boldsymbol{B}$. The Nyströmformer approximates the $\mathrm{Softmax}$ matrix as:
\begin{equation}
  \mathrm{Softmax} \left( \frac{\boldsymbol{Q} \boldsymbol{K}^\top}{\sqrt{d}} \right) \approx \boldsymbol{S}_{Q\tilde{K}} \boldsymbol{S}_{\tilde{Q}\tilde{K}}^+ \boldsymbol{S}_{\tilde{Q}K}
\end{equation}
\noindent where the superscript $+$ denotes the Moore-Penrose inverse typically computed with the singular value decomposition (SVD). Since the SVD is inefficient on GPU, the authors relied on an iterative method that approximate $\boldsymbol{S}_{\tilde{Q}\tilde{K}}^+$ as $Z^+$. Finally, the Nyströmformer's attention is given by:
\begin{equation}
  \mathrm{Attention}(\boldsymbol{X}) \approx \boldsymbol{S}_{Q\tilde{K}} Z^+ \boldsymbol{S}_{\tilde{Q}K} \boldsymbol{V}
\end{equation}
\noindent which can be efficiently encoded in a computational graph.

Provided that the number of landmarks is constant and much smaller than the sequence length, the Nyströmformer complexity is $\mathcal{O}(n)$. Depending on the number of landmarks and the sequence length, the authors reported substantial gains over the Linformer and Longformer on the masked language model and sentence order prediction objectives. Additionally, the representations learned by the Nyströmformer appear to transfer as well as BERT to different NLP tasks. Nonetheless, a more extensive evaluation of the Nyströmformer remains necessary.

\textbf{Kernel Attention}~\citep{choromanski2020rethinking, katharopoulos2020transformers}: A kernel $K(\cdot, \cdot)$ is a function that takes two vectors as arguments and returns the product of their projection by a feature map $\phi(\cdot)$:
\begin{equation}
  K(\boldsymbol{x}, \boldsymbol{y}) = \phi(\boldsymbol{x})^\top\phi(\boldsymbol{y})
\end{equation}
\citet{katharopoulos2020transformers} interpreted the $\mathrm{Softmax}$ as a kernel, decomposed it as an inner product in the right space, and rearrange the computations in a clever way to reduce the complexity. More specifically, the self-attention of a given query $\boldsymbol{Q}_i$ may be rewritten using a mapping $\phi(\cdot)$:
\begin{equation}
    \resizebox{0.92\linewidth}{!}{%
    $\mathrm{Softmax}\big(\boldsymbol{Q}_i^\top \boldsymbol{K}^\top\big)\boldsymbol{V}  = \frac{\sum_{j=1}^n \mathrm{exp}\big(\boldsymbol{Q}_i^\top \boldsymbol{K}_j\big) \boldsymbol{V}_j}{\sum_{j=1}^n \mathrm{exp}\big(\boldsymbol{Q}_i^\top \boldsymbol{K}_j\big)}                  
    = \frac{\sum_{j=1}^n \phi\big(\boldsymbol{Q}_i\big)^\top \phi\big(\boldsymbol{K}_j\big) \boldsymbol{V}_j}{\sum_{j=1}^n \phi\big(\boldsymbol{Q}_i\big)^\top \phi\big(\boldsymbol{K}_j\big)}      
    = \frac{\phi\big(\boldsymbol{Q}_i\big)^\top \sum_{j=1}^n \phi\big(\boldsymbol{K}_j\big) \boldsymbol{V}_j^\top}{\phi\big(\boldsymbol{Q}_i\big)^\top \sum_{j=1}^n \phi\big(\boldsymbol{K}_j\big)}$%
    }
\end{equation}
\noindent where the scaling factor $\sqrt{d}$ has been omitted for the sake of readability. The authors noted that $\sum_{j=1}^n \phi\big(\boldsymbol{K}_j\big) \boldsymbol{V}_j^\top$ and $\sum_{j=1}^n \phi\big(\boldsymbol{K}_j\big)$ must only be computed a single time, therefore reducing the complexity from quadratic to linear both in terms of memory and computation. The vectorized formulation of the numerator makes it simpler to see:
\begin{equation}
  \underbrace{\vphantom{\big(\boldsymbol{Q}^\top}\phi\big(\boldsymbol{Q}\big)}_{n \times p} \Big(\underbrace{\phi\big(\boldsymbol{K}\big)^\top}_{p \times n} \underbrace{\vphantom{\big(\boldsymbol{Q}^\top}\boldsymbol{V}}_{n \times d}\big)
\end{equation}
\noindent where the mapping $\phi(\cdot):\mathbb{R}^{d} \rightarrow \mathbb{R}^{p}$ is applied position-wise. Unfortunately, the feature map of the exponential kernel is infinite dimensional. Hence, any finite kernel is an approximation of the attention matrix and may be interpreted as a low-rank factorization. However, they are presented separately here due to their conceptual difference. \citet{katharopoulos2020transformers} approximated the attention matrix in the Linear Transformer with the feature map $\phi(x)=\mathrm{elu}(x)+1$, where the function $\mathrm{elu}(\cdot)$ denotes the exponential linear unit given by:
\begin{equation}
  \mathrm{elu}(x) = \left\{
  \begin{array}{ll}
    \alpha (e^x -1), & x < 0   \\
    x,               & x\geq 0 \\
  \end{array}
  \right.
\end{equation}
\noindent where $\alpha$ is an hyperparameter. The Linear Transformer performed on par with the vanilla Transformer on autoregressive image generation, but poorly on automatic speech recognition.

\citet{choromanski2020rethinking} later demonstrated that the exponential is equivalent to a kernel with a randomized mapping:
\begin{equation}
    \mathrm{exp}(x^\top y)=
    \mathbb{E}_{w\sim \mathcal{N}(0,I_d)}\bigg[\mathrm{exp}\left(w^\top x\frac{\lVert x\lVert ^2}{2}\right)\mathrm{exp}\left(w^\top y\frac{\lVert y\lVert ^2}{2}\right) \bigg]
\end{equation}
Consequently, the authors introduced the Performer, a linear complexity model that approximates the attention by means of a kernel with the following feature mapping:
\begin{equation}
    \phi(x)=\frac{\mathrm{exp}(-\lVert x\lVert ^2/2)}{\sqrt{2p}}\big[\mathrm{exp}\big(w_1^\top x\big);\dots;\mathrm{exp}\big(w_p^\top x\big);\mathrm{exp}\big(-w_1^\top x\big);\dots;\mathrm{exp}\big(-w_p^\top x\big)\big]
\end{equation}
\noindent where $w_i\sim \mathcal{N}(0, I_d)$. To further reduce the variance of the estimator, $w_i$ are constrained to be exactly orthogonal, which is achieved with the Gram-Schmidt process. The hyperparameter $p$ corresponds to the number of random features and controls the quality of the approximation.

\textbf{Clustering and Locality-Sensitive Hashing}~\citep{vyas2020fast}: As previously explained, clustering can uncover sparse patterns by grouping queries and keys and only computing the attention between positions within the same cluster. Alternatively, \citet{vyas2020fast} proposed to factorize the attention with clustering by grouping queries into a fixed number of non-overlapping clusters and by computing the attention between the cluster's centroids and the keys. Consequently, the attention score is only computed once per group of similar queries and broadcasted to all, resulting in linear complexity. Since queries may be clustered differently across attention heads and since the attention sub-layer includes a residual connection, two queries in the same cluster can have different output representations. The authors proved that the approximation error for a given query is bounded by its distance to its centroid multiplied by the spectral norm of the keys matrix. As such, the K-Means algorithm can be used for minimizing the approximation error. However, K-Means in the original space would be slow to compute as Lloyd algorithm has a complexity of $\mathcal{O}(ncdl)$, where $c$ is the number of clusters and $l$ is the number of Lloyd iterations. Instead, the authors first used a locality-sensitive hashing scheme on the queries before applying K-Means with the Hamming distance, which reduces the complexity to $\mathcal{O}(ncl + cbl + ndb)$, where $b$ is the number of bits used for hashing.

To further improve the approximation, \citet{vyas2020fast} proposed the \textit{improved cluster attention} that separately consider the $k$ keys with the highest attention for each cluster. Intuitively, keys with high approximated attention may have low attention for some queries, resulting in a large approximation error. As a solution, the dot product between these top-$k$ keys and all queries belonging to the corresponding cluster is computed. Then, the attention is rescaled by the total probability mass assigned to these top-$k$ keys.

Compared to the Reformer, \citet{vyas2020fast} method is significantly faster ($43\%$ lower epoch time) while being significantly more accurate ($35\%$ lower phone error rate) for speech recognition on the Wall Street Journal.

\subsection{Architectural Change}
\label{ssec:architectural}

Finally, the Transformer's complexity may also be reduced by modifying the model's architecture and preserving the original attention mechanism. Let us investigate (i) the Transformer-XL and the Compressive Transformer that rely on memory, and (ii) then the Funnel-Transformer that iteratively compresses sequences.

\begin{figure}[htb]
  \centering
  \Description[Framework of the Transformer-XL.]{Framework of the Transformer-XL.}
  \includegraphics[scale=0.47]{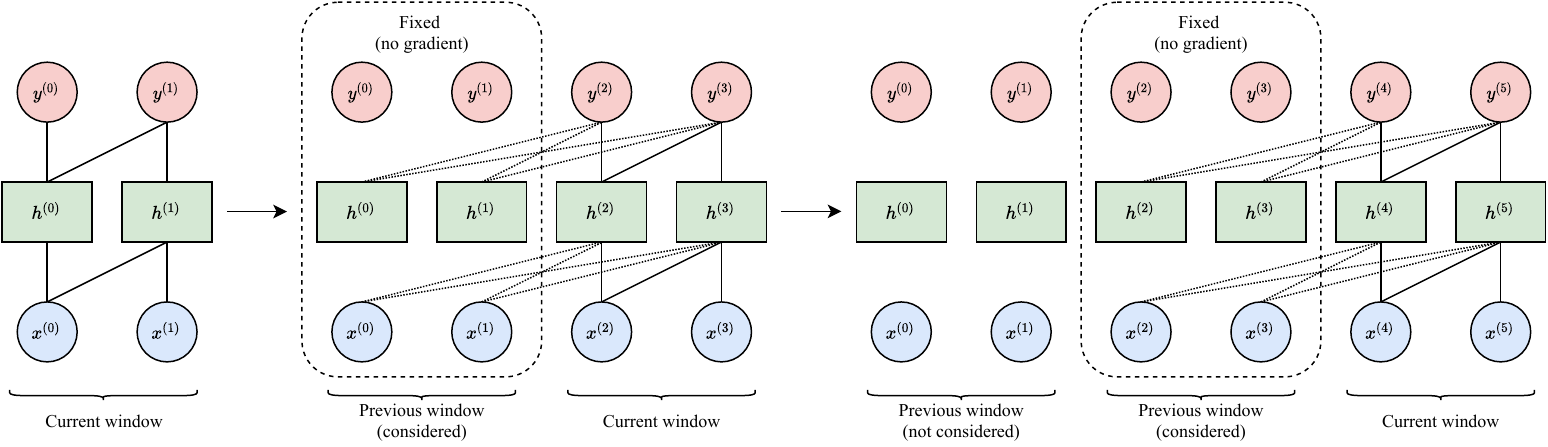}
  \caption{Segment-based recurrence, which is similar to truncated BPTT. The window size is equal to two, and only the previous window is considered. For the sake of clarity, parameters from and to states that do not contribute are omitted.}
  \label{fig:transformerxl}
\end{figure}

\textbf{Memory}~\citep{dai-etal-2019-transformer, DBLP:conf/iclr/RaePJHL20}: The block-wise approach splits the input sequence into small non-overlapping subsequences called windows, blocks, or chunks, which are processed independently; therefore, the maximum dependency length is equal to that of the subsequence. To leverage information from previous windows, \citet{dai-etal-2019-transformer} introduced the Transformer-XL, which relies on segment-based recurrence between windows.  This recurrence mechanism is implemented by storing the representations of the previous window in a first-in first-out memory (FIFO). Then, the attention mechanism can attend to the representations located in this memory, but the gradients are not computed for the attention on these elements. Although this model achieves a RECL four times greater than the vanilla Transformer with the same parameter budget, it cannot capture dependencies outside the FIFO memory range. Furthermore, this model is only compatible with autoregressive tasks. This technique is analogous to truncated backpropagation through time (BPTT), except that a sequence of hidden states is considered instead of the previous one. Figure~\ref{fig:transformerxl} illustrates the segment-based recurrence of the Transformer-XL.

 In order to further increase the range of dependencies considered by the Transformer-XL, \citet{DBLP:conf/iclr/RaePJHL20} proposed the Compressive Transformer, which adds a compressed memory to the original FIFO memory. Representations of past windows are first stored in the standard FIFO memory, like the Transformer-XL. Then, when this memory is full, the oldest representations are compressed with a user-defined function and stored in the compressed FIFO memory instead of being discarded. The number of elements considered in the original FIFO memory to generate the compressed memory depends on the chosen function. The authors propose using max/mean pooling, 1D convolution, dilated convolutions, or the most attended representations by the attention. They also proposed to learn the compression function with an auxiliary auto-encoding loss and a variant called attention-reconstruction loss, which typically reconstructs the original memory from the compressed ones. They show a clear advantage over the Transformer-XL on NLP tasks and comparable results on speech modelling.

\textbf{Sequence Compression}~\citep{dai2020funneltransformer}: Many tasks such as image classification and sentiment analysis only require producing a single output for the whole sequence. \citet{dai2020funneltransformer} argued that the full-length sequence of hidden states may contain significant redundancy and that the model may not have to preserve token-level information. Consequently, they proposed the Funnel-Transformer, whose encoder reduces the computational cost by gradually reducing the length of the hidden states sequence with pooling. Note that instead of directly feeding the pooled sequence into the attention layer, it is only used to construct the query matrix, while the unpooled sequence is used to construct the key and value matrices. Additionally, the authors proposed to recover the original sequence length by up-sampling the compressed sequence of hidden states to address the common pre-training objectives, such as MLM, that require separate representation for each token. Although the Funnel-Transformer effectively reduces the computational and memory cost of the encoder, the complexity remains quadratic, and the best performances are achieved on tasks that only require sequence-level representation.

\section{Shortcomings}
\label{sec:shortcomings}

This section discusses the lack of understanding of the self-attention inner workings and the limitation of the Transformer evaluation methodology, including the lack of standard benchmarks for long-range dependencies.

Self-attention is a relatively new mechanism that has been quickly and widely adopted due to its remarkable empirical success. Nonetheless, the self-attention inner workings are not yet fully understood, and many questions remain unanswered, including why it works, what it learns, and whether it is interpretable. Answering those questions is crucial to designing faster and lighter Transformers that are competitive with the original model. As of this paper's writing, the deep learning community actively investigates self-attention and have proposed preliminary answers to the aforementioned questions. For instance, evidence supporting both the self-attention interpretability~\citep{wiegreffe2019attention, serrano-smith-2019-attention} and non-interpretability~\citep{jain2019attention} have been published. \citet{tay2020synthesizer} empirically evaluated the dot product impact on natural language processing tasks and concluded that query-keys interaction is ``\textit{useful but not that important}''. \citet{Kitaev2020Reformer} investigated the impact of sharing queries and keys, and concluded that ``\textit{it turns out that sharing QK does not affect the performance of Transformer}''.

Despite our current limited understanding of the self-attention mechanism, a wide range of faster and lighter Transformers have been introduced in a short amount of time, each claiming comparable or superior performance to the vanilla Transformer. Since there is no consensus on how to evaluate the proposed approaches~\citep{tay2020long}, researchers often have to evaluate their method on a small range of tasks. However, different tasks may require different assumptions, which means that one method may work well on a specific task but poorly on others. For instance, \citet{tay2020synthesizer} showed that a simple Synthesizer is highly competitive with the vanilla Transformer across a range of natural language processing tasks, including machine translation, language modelling, and text generation. However, \citet{tay2020long} later showed that the vanilla Transformer outperforms the Synthesizer on the more difficult Long-Range Arena benchmark. Long-Range Arena~\citep{tay2020long} is a suite of five general and challenging tasks designed to evaluate how well Transformers capture long-term dependencies from different modalities such as text, natural and synthetic images, and mathematical expressions. Table~\ref{tab:benchmark} compiles the Long-Range Arena results of the models discussed in the survey. For a complete description of the objectives and datasets, we refer the reader to the original paper.

Furthermore, due to Transformers large training cost, researchers often evaluate their approach against a limited number of models on the tasks of interest. For instance, \cite{Kitaev2020Reformer} only evaluated the Reformer against three distinct vanilla Transformers~\citep{NIPS2017_7181, ott2018scaling} on three tasks. Standardized suites of benchmarks such as GLUE and the recent Long-Range Arena allow researchers and practitioners to evaluate only their method and compare it against a public leaderboard. Consequently, we highly recommend that researchers consider such benchmarks.

Although standardized benchmarks such as Long-Range Arena would help compare the models, the results should be taken with caution since the performance depends on the model size and hyperparameters, the speed depends on the implementation and hardware, and the memory footprint depends on the implementation and general methods used. For instance, the Switch Transformer uses a mixture of experts, mixed-precision, expert dropout, knowledge distillation, and a careful initialization. Therefore, it is difficult to isolate the benefit of a single modification.

Finally, the complexity is not always representative of the practical efficiency. For instance, the Reformer achieves an asymptotic complexity of $\mathcal{O}(n \log n)$ but is significantly slower than the vanilla Transformer on small sequences, as shown in Table~\ref{tab:benchmark}. This slow down is due to large constants hidden in the complexity. Even when there are no hidden constants, there is a distinction between theoretical complexity and what is achievable in practice. For instance, sparse matrix multiplication may reduce the complexity from quadratic to linear in theory. However, it is well known that GPUs and TPUs are not designed to perform such operations efficiently~\citep{4625887} and, in practice, sparse matrix multiplication is often slower than dense ones. We encourage researchers to explicitly report the complexity as well as the number of floating operations (FLOPs), the wall-clock time with the hardware, and the memory footprint of their method.

\begin{table}[htb]
  \centering
  \small
  \caption{Long-Range Arena benchmark~\citep{tay2020long}. Results have been compiled from the original paper. Benchmarks are run on 4x4 TPU V3 chips, and the memory is reported per device.}
  \begin{tabular}{lccccc}
    \multirow{2}{*}{Models}                                                                     & \multirow{2}{*}{Average score (\%)} & \multicolumn{2}{c}{Steps per second} & \multicolumn{2}{c}{Peak memory (GB)}                                 \\
                                                                                                &                                     & 1K                                   & 4K                                   & 1K            & 4K            \\ \hline
    Transformer~\citep{NIPS2017_7181}                            & 54.39                               & 8.1                                  & 1.4                                  & 0.85          & 9.48          \\ \hline
    Sparse Transformer\footnotemark[\value{footnote}]~\citep{DBLP:journals/corr/abs-1904-10509} & 51.24                               &                                      &                                      &               &               \\
    Longformer\footnotemark[\value{footnote}]~\citep{2020arXiv200405150B}                                                      & 53.46                               &                                      &                                      &               &               \\
    BigBird~\citep{NEURIPS2020_c8512d14}                                                        & \textbf{55.01}                      & 7.4                                  & 1.5                                  & 0.77          & 2.88          \\
    Sinkhorn Transformer~\citep{tay2020sparse}                                                  & 51.39                               & 9.1                                  & 5.3                                  & 0.47          & 1.48          \\
    Reformer~\citep{Kitaev2020Reformer}                                                         & 50.67                               & 4.4                                  & 1.1                                  & 0.48          & 2.28          \\
    Linformer~\citep{2020arXiv200604768W}                                                       & 51.36                               & 9.3                                  & 7.7                                  & \textbf{0.37} & \textbf{0.99} \\
    Synthesizer~\citep{tay2020synthesizer}                                                      & 51.39                               & 8.7                                  & 1.9                                  & 0.65          & 6.99          \\
    Linear Transformer~\citep{katharopoulos2020transformers}                                    & 50.55                               & 9.1                                  & 7.8                                  & \textbf{0.37} & 1.03          \\
    Performer~\citep{choromanski2020rethinking}                                                 & 51.41                               & \textbf{9.5}                         & \textbf{8.0}                         & \textbf{0.37} & 1.06          \\ \hline
  \end{tabular}
  \label{tab:benchmark}
\end{table}
\footnotetext[\value{footnote}]{The Sparse Transformer and Longformer depends on CUDA kernels that are difficult to implement on TPUs. Therefore, \citet{tay2020long} used \textit{equivalent} implementations to emulate their performance and did not report their efficiency.}

\section{Broader Impact of Efficient Transformer}
\label{sec:impact}

This section extends the three motivations and potential impacts of lighter and faster Transformers briefly discussed in Section~\ref{ssec:complexity}.

First and foremost, computational resources are not only finite but also expensive. Consequently, there are severe inequalities between research groups and between companies. Indeed, many researchers do not have access to GPU or TPU farms, and most companies cannot afford to spend thousands or millions of dollars on dedicated hardware, especially if deep learning is not their primary focus. At this time, the resources disparities have increased dramatically to a point where only a few parties can afford to train massive state-of-the-art models. A prime example of this cleavage is the Transformer. Indeed, the largest Transformers are so expensive to train, even for large companies such as Microsoft, that they are only trained once. For instance, \citet{brown2020language} noticed an issue in their pre-processing after training GPT-3. As the author explained, they could not train their model again due to the massive cost and therefore published their results with a known issue. Resources inequalities also hinder creativity as researchers with promising ideas may not be able to implement them, thus reinforcing the vicious ``rich get richer'' circle, where well-funded groups and companies that have access to more resources are more likely to achieve state-of-the-art results and receive more fundings~\cite{strubell2019energy}.

Additionally, lower-complexity Transformers enable novel applications as extremely long sequences cannot be processed in a reasonable amount of time by the quadratic complexity vanilla Transformer. For instance, \citet{choromanski2020rethinking} observed the Performer's potential impact on biology, and \citet{NEURIPS2020_c8512d14} evaluated BigBird on genomics tasks that take fragments of DNA as input. \citet{47717} were able to generate minute-long musical compositions with a Transformer that leverage the block-wise approach and an efficient computation of the relative attention. Note that contrary to the attention introduced by \cite{NIPS2017_7181}, the relative attention~\citep{shaw2018selfattention} explicitly models the input positions. The range of applications will surely expand as researchers design ever-lighter and -faster Transformers.

Finally, recent research made it clear that we must cut carbon dioxide (CO2) emissions in half over the next decade to limit global warming. The large-scale infrastructures used by the deep learning community consume a considerable amount of electricity, which is mainly produced by non-renewable sources such as coal or gas~\citep{iea}. \citet{strubell2019energy} estimated that training a Transformer with neural architecture search generates up to 284,000 kg of CO2. For reference, the average American emits 16,400 kg of CO2 per year, and the average car emits about 57,200 kg during its lifetime\footnote{A product lifetime or lifecycle typically includes material production, manufacturing, usage, and end-of-life disposal.} (fuel included). The authors estimated that training a single instance of BERT~\citep{devlin2019bert} on GPU produces about the same amount of CO2 as a trans-American flight. Although lighter and faster models require fewer resources and therefore produce less carbon dioxide, they are also more accessible, so we would expect more models to be trained. Overall, it is difficult to know whether lighter and faster Transformers will positively impact the environment. Nonetheless, researchers and practitioners ought to have in mind the significant environmental impact of their experiments, which can be estimated with the Machine Learning Emissions Calculator\footnote{\url{https://mlco2.github.io/impact/}} developed by \citet{9123463}.

\section{Future Research Directions}
\label{sec:directions}

In our opinion, the current research directions follow one of two purposes: (i) efficiency and affordability or (ii) generalization performance. Since this survey addresses approaches to yield faster and lighter Transformers, let us start with the efficiency and affordability objective.

\subsection{Efficiency and Affordability}

To the best of our knowledge, researchers and practitioners have not yet identified a specialized approach that improves the Transformer's efficiency for every task, dataset, and hardware, as explained in Section~\ref{sec:shortcomings}. In our opinion, one of the most promising avenues is to learn adaptively sparse patterns that are structured for the available hardware. Let us justify our claim. 

The $\mathrm{Softmax}$ function only contains a few large values due to its exponential nature. Therefore, it can be effectively approximated by masking the positions with small weights. In theory, the computation and memory reduction is linearly proportional to the ratio of masked positions. In practice, however, the improvement depends on the hardware. As of this survey's writing, NVIDIA is the first and only manufacturer to offer an architecture that natively supports sparse operations, resulting in a virtually perfect speed-up. One may reasonably expect other manufacturers to follow this direction due to the prevalence of sparse operations in deep learning. Therefore, the sparse patterns should be structured such that the hardware natively supports them. Handcrafting features or patterns based on prior knowledge is known to be suboptimal. Instead, the model should learn the patterns from the data for the task at hand. Additionally, individual samples are likely to require different attention patterns, and hence, the patterns should be adaptative (content-based). Finally, we believe it is beneficial to include global tokens since they allow any position to attend to any other position in two layers, thus preserving the attention's expressiveness.

\subsection{Generalization Performance}

A second research venue consists in improving the network generalization performance. Since the deep learning renaissance associated with greedy layer-wise unsupervised pre-training~\citep{Goodfellow-et-al-2016}, there has been a clear trend towards scaling up neural networks. As a result, researchers and practitioners have been able to leverage ever-larger datasets and ultimately improve the network's performance. In this setting, scaling is performed typically by increasing the number of layers, the number of attention heads, the input embedding dimension, and the feedforward network width.

Amongst others, \citet{gpt-2} introduced a large Transformer called GPT-2 and evaluated various model sizes on language modelling tasks in a zero-shot setting. The authors reported that the performance significantly increased with the model size ranging from 117M to 1.5B parameters. Recently, \citet{brown2020language} introduced GPT-3 based on the GPT-2 architecture and considered an even wider span of model sizes, ranging from 125M to 175B parameters. The authors reported that the model performance smoothly increased with the model size in most cases and suggested that this trend should extend to even larger models. Furthermore, \citet{devlin2019bert} investigated the effect of BERT size on the GLUE benchmark and concluded that ``\textit{larger models lead to a strict accuracy improvement across all four datasets, even for MRPC which only has 3,600 labeled training examples, and is substantially different from the pre-training tasks}''.

These observations suggest that researchers and practitioners must scale their model to pursue the generalization performance objective. Inherently, scaling is resource-expensive and goes against the affordability sought in this survey. Nonetheless, there are research directions to improve the generalization capability of deep learning models that are orthogonal to scaling and thus compatible with efficiency. A promising avenue is structural inductive biases. A recent structural inductive bias inspired by independent mechanisms in the causality literature consists of designing an architecture that learns sparsely interacting modules, each one of them specialized in a different mechanism~\citep{goyal2020recurrent}. Ideally, individual modules should be robust to changes in the aspects of the world that are unrelated to this module, such as in the case of distributional shift. \citet{lamb2021transformers} applied this idea to Transformers by introducing the Transformers with Independent Mechanisms (TIM). The authors observed that TIM layers could be combined with the mixture of experts approach, allowing the switching to be specific to distinct aspects of the data.

Combining universally effective and efficient approaches such as the aforementioned sparse patterns with conditional computing and the independent mechanisms prior appears to be promising to tackle complex tasks without relying on large-scale resources.

\section{Conclusion}
\label{sec:conclusion}

Transformers have quickly become the de facto model for processing sequences, notably achieving state-of-the-art in most natural language processing tasks at the cost of quadratic complexity. As a result, researchers have leveraged numerous techniques to mitigate this memory and computational burden. This survey investigated popular general methods to make neural networks lighter and faster and discussed their strengths and limitations. Notably, we advised researchers and practitioners to use mixed-precision and gradient checkpointing due to their simplicity and overall benefits. Often, these general techniques are not sufficient to mitigate the Transformer's complexity. Consequently, this survey reviewed the lower-complexity variations of the Transformer and discussed their assumptions, justification and shortcomings. Notably, we advised researchers and practitioners to rely on pre-trained models whenever possible. Otherwise, we recommend training a small vanilla Transformer with mixed-precision and gradient checkpointing to apprehend the dependencies required for the task and select the appropriate models accordingly. Additionally, we discussed the potential impacts of affordable Transformers, including improving the state-of-the-art, extending the range of applications, increasing the equity between researchers, and potentially reducing the environmental impact. Finally, we highlighted promising future research directions for this exciting architecture.

\begin{acks}
We would like to gratefully acknowledge the Natural Sciences and Engineering Research Council of Canada (NSERC), Prompt, Ericsson, Ciena, and EfficiOS for funding this research.
\end{acks}

\input{bibliography.bbl}

\newpage
\appendix

\section{Introduction to Machine Learning}
\label{sec:introduction_to_ml}

At the dawn of artificial intelligence (AI), researchers rapidly tackled and solved problems that were challenging for humans but relatively straightforward for computers in that they could be described as a set of rules. Chess may certainly be the epitome of such complex tasks solved brilliantly by artificial intelligence. Nonetheless, despite the achievements of AI, simpler tasks that humans solve instinctively proved to be much more challenging as they are not easily expressed formally. Amongst others, speech and object recognition were -- and still are to some extent -- challenging problems to solve for computers. Machine learning (ML) provides a solution to intuitive problems by allowing computers to learn from experience instead of relying on human knowledge to specify the tasks or their solution. The seemingly ever-increasing amount of data produced every day has enabled machine learning to become suitable and successful for a wide range of simple and complex problems. Since the renaissance of deep learning (DL) associated with greedy layer-wise pre-training, neural networks have become the most popular family of algorithms for machine learning as they learn a hierarchy of concepts, with each concept defined through its relation to simpler concepts. As of the writing of this survey, deep learning, and more generally artificial intelligence, has become a thriving field with numerous practical applications that directly impact countless human lives, from medical diagnoses to movie recommendations.

\section{Practical Guidelines - General Methods}
\label{sec:guidelines_generic}

The general approaches presented in Section~\ref{sec:generic} apply to the original Transformer as well as its lower-complexity alternatives. Therefore, they are discussed before introducing the specialized approaches in Section~\ref{sec:specialized}. In particular, this section provides practitioners and researchers with a series of guidelines on which methods to apply depending on the bottleneck and whether it occurs during optimization or inference. The distinction between optimization (i.e. pre-training and training) and inference is motivated by the former being significantly more resource-intensive than the latter.

The primary focus of this survey is to make Transformers more efficient and ultimately more affordable. Therefore, only substantial performance losses will be mentioned, along with other significant drawbacks such as incompatibilities and instabilities. Unless specified otherwise, the methods are readily available in PyTorch~\citep{NEURIPS2019_9015} and Tensorflow~\citep{tensorflow2015-whitepaper}, two standard deep-learning libraries.

\subsection{Optimization}

Optimization is the most resource-intensive phase, prominently due to the iterative nature of the process, the quadratic complexity of the attention mechanism, and the in-memory recording of intermediate values during the forward pass. Consequently, most of the above approaches to reduce computation, memory, or both, focus on optimization.

\subsubsection{Computation Savings}

Recently, the undeniable success of pre-trained Transformers such as BERT~\citep{devlin2019bert}, ViT~\citep{dosovitskiy2020image}, and GPT-3~\citep{brown2020language} has confirmed the benefits of unsupervised pre-training. As previously mentioned, pre-training initializes the network's weights in a ``good'' region of space that allows the model to converge faster. Therefore, we advise practitioners and researchers to build upon pre-trained models like the ones available on the open-source library Hugging Face~\citep{wolf-etal-2020-transformers}.

Nonetheless, pre-trained models are typically only available for ``conventional'' data and tasks such as translation, summarization, question answering, text-to-speech, image classification, and object detection. As for data and tasks without pre-trained models, we recommend initializing the model with a principled strategy such as Admin or T-Fixup and using a sample-efficient objective. Those techniques are not yet implemented in standard libraries, therfore we suggest using T-Fixup as it is simpler than Admin.

\subsubsection{Memory Savings}

As discussed before, although time may limit one's experiments, memory bottlenecks are much more critical. Since the intermediates values are responsible for a substantial part of the memory footprint, the first method to apply whenever memory is the main limiting factor during optimization is gradient checkpointing. The approach has two significant advantages: (i) the trade-off between memory and computation controlled by the number of intermediate values kept in memory is highly adjustable, and (ii) the method is straightforward to use in  TensorFlow\footnote{\url{https://github.com/cybertronai/gradient-checkpointing}} and PyTorch\footnote{\url{https://pytorch.org/docs/stable/checkpoint.html}}. Nevertheless, gradient checkpointing has some caveats with multiple GPUs, even on a single machine. For instance, as of the writing of this survey, gradient checkpointing interferes with PyTorch's Distributed and Data Parallel API, leading to instabilities\footnote{\url{https://discuss.pytorch.org/t/ddp-and-gradient-checkpointing/132244/2}}.

Alternatively to gradient checkpointing, reversible layers provide a mechanism to recompute the intermediate values during the backward pass, thereby decoupling the model's depth from the amount of memory required by the activations. Although the increase in computation is reasonable, reversible layers produce numerical errors that may accumulate layers after layers to the point that they become an issue. Additionally, reversible layers are not yet part of standard libraries and require manually writing the forward and backward operations.

In addition to gradient checkpointing or reversible layers, parameter sharing allows further reducing the memory and is straightforward. However, unlike the other approaches, parameter sharing reduces the model's capacity. Fortunately, the trade-off between capacity and memory/computation savings is highly customizable, depending on the number of parameters shared.

Finally, a mixture of experts potentially with micro batching is expected to allow many memory-limited GPUs to train a Transformer even if each GPU is individually too small. However, both approaches require substantial effort to implement and impose a communication cost.

\subsection{Inference}

Sometimes, researchers have the resources to train large models during the development phase due to public or academic infrastructures, but they do not have the resources to deploy them. In such cases, one may do a neural architecture search to find the best model within a parameter budget during training, preferably with \citet{SoLL19}'s approach. As of this survey's writing, neural architecture search is not part of standard libraries.

Alternatively or additionally to NAS, structured pruning and distillation reduce the amount of memory and computations with fine-grained control. While structured pruning is already implemented, distillation is as easy as building a second model that predicts the teacher's output. As the aforementioned results suggest~\citep{2020arXiv200403844S,michel2019sixteen,sanh2020distilbert,tsai2019small,jiao2020tinybert}, the Transformer's performance does not significantly degrade when the model is pruned or distilled. Therefore, to reduce the amount of energy consumed by the model, we suggest applying those methods even when resources are sufficient during inference.

\subsection{Optimization and Inference}

The first and foremost method for faster and lighter models is automatic mixed-precision. Mixed-precision is compatible with virtually every neural network, combines with every other approach, reduces the memory footprint and accelerates computations on modern GPUs. Additionally, this method is one of the simplest to implement, only requiring a few lines of code in PyTorch\footnote{\url{https://pytorch.org/tutorials/recipes/recipes/amp_recipe.html}} and TensorFlow\footnote{\url{https://www.tensorflow.org/guide/mixed_precision}}.

Although 8-bit quantization may seem similar to 16-bit mixed-precision, the former is primarily used to speed up inference and is not as readily available as the latter. In particular, PyTorch does not provide quantized operators for GPU as of the writing of this survey, and Tensorflow warns users that ``\emph{different hardware may have preferences and restrictions that may cause slight deviations when implementing the spec that result in implementations that are not bit-exact}''\footnote{\url{https://www.tensorflow.org/lite/performance/quantization_spec\#specification_summary}}. Due to the finicky nature of 8-bit quantization, we suggest reserving this approach to specific hardware and use-cases such as the mobile setting.

\section{Practical Guidelines - Specialized Methods}
\label{sec:guidelines_specialized}

With limitations mentioned in Section~\ref{sec:shortcomings} in mind, let us examine the results of \citep{tay2020long} and draw some broad guidelines.

The first observation is that every model is lighter than the original Transformer. Nonetheless, for memory-limited environments, the Synthesizer is the least advisable alternative as the model only reduces the memory by 24 to 26\% regardless of the sequence length, which is consistent with its quadratic complexity. Instead, the Linformer, Performer, and Linear Transformer are better suited to address memory bottlenecks as they are at least 56\% and 88\% lighter than the original Transformer for input sequences of 1,000 and 4,000 tokens, respectively, which is also consistent with their linear complexity.

The second observation is that, on TPU V3 chips, the Synthesizer, the Reformer and BigBird perform roughly the same number of steps per second as the original Transformer regardless of the sequence length. In contrast, the Linformer, Performer, Sinkhorn Transformer and Linear Transformer are significantly faster than the original Transformer for input sequences of 4,000 tokens while performing on par for sequences of 1,000 tokens. Consequently, those models are better suited for computation-limited environments. We do not wish to overstate our claims here since TPUs and GPUs differ on some key aspects\footnote{Compared to modern GPUs with Tensor Cores, TPUs typically perform more FLOPs but have a lower memory bandwidth, have fewer but larger tiles, and apply the activation function within the matrix multiplication.}, and speed-ups may significantly vary, as observed by \citet{wang2019benchmarking} and \citet{9139681}. Although the data processing pipeline and the model implementation are outside this survey's scope, they should be tuned for the exact hardware used as it may significantly impact the performance.

Nonetheless, it would seem that the Linformer, Performer, and Linear Transformer are excellent options to improve memory and computation, with the Linformer standing out considering the simplicity of its implementation. However, those models also have serious drawbacks. The Linformer requires instantiating the projection matrices $\boldsymbol{E}$ and $\boldsymbol{F}$ of dimension $k \times n$, and thus can only process fixed-sized input sequences. Therefore, sequences must be padded to the size of the largest one in the dataset, which may significantly degrade the model's efficiency. The Performer and Linear Transformer are challenging to be  efficiently implemented. Besides, they perform noticeably worse than the original Transformer on average. In some cases, such as byte-level text classification, they manage to outperform the original Transformer. In other cases, however, they might critically underperform. For instance, in a longer variant of the ListOps task~\citep{Nangia2018ListOpsAD} that consist of modelling hierarchically structured data, they achieve less than 50\% of the original Transformer's performance.

In contrast, sparse Transformers suffer less performance degradation on average, as measured on the Long-Range Arena benchmark. Notably, the LongFormer and BigBird achieved the same accuracy as the original Transformer for the ListOps task. Sparse models have, however, two major shortcomings. First, the sparsity must be structured in order to be efficiently implemented and yield practical improvements. Otherwise, the sparse model may be slower than its dense equivalent. Furthermore, CUDA kernels require considerable effort to be efficiently implemented and are specific to GPUs. Implementing equivalent kernels on TPUs is challenging, or even impossible, due to the disparity in supported primitives and operations. Secondly, dependencies that must be modelled to solve the task accurately should not be masked. Otherwise, the performance will be critically impacted. To select the appropriate sparse model, we recommend that one train a small vanilla Transformer with mixed-precision and gradient checkpointing,  and then analyze the activation patterns of each layer's attention.

Nonetheless, in a recent paper, \citet{narang2021transformer} investigated the impact of numerous modifications to the Transformer architecture, including changes in activation, normalization, depth, embeddings, and Softmax, on three NLP benchmarks, namely SuperGLUE~\citep{wang2020superglue}, XSum~\citep{narayan2018dont}, and WebQ~\citep{berant-etal-2013-semantic}. The authors also evaluated several methods studied in this paper, including parameter sharing, Synthesizers, the Switch Transformer, and the Universal Transformer. They observed that no modification was able to improve the performance significantly. After ruling out various explanations, the authors conjectured that ``\textit{modifications to the Transformer architecture often do not transfer across implementations and applications}'', which may explain why no modification has been widely adopted.

In conclusion, there seem to be no simple and universal guidelines regarding the current Transformer alternatives. If the data and task are standard, we recommend looking in the literature or on the \href{https://paperswithcode.com}{Papers With Code} website for references on how the different methods compare and experiment with already pre-trained models. Otherwise, we recommend using a small vanilla Transformer with mixed-precision and gradient checkpointing as baseline, then experimenting with already implemented lower-complexity models. As a side note, one may also want to combine multiple specialized approaches. For instance, BigBird-ETC relies on additional tokens for global attention, a form of memory similar to the Compressive Transformer. Nonetheless, many combinations are unprincipled at best. For instance, one should not factorize a sparse attention: the complexity will be similar to that of the same factorization of the full attention, and the sparsity may lose valuable information that the factorization could have preserved.

\section{Alternatives to Self-Attention}
\label{sec:alternative}

Recently, attention-free alternatives to the Transformer have been proposed, putting \citet{NIPS2017_7181} original paper title \textit{Attention Is All You Need} to the test. Such architectures have not been explored in the core of this survey as they arguably remove the Transformer's core mechanism. Nonetheless, it is important to mention some of the most popular and promising alternatives.

\citet{tolstikhin2021mlpmixer} argued that self-attention is not required for image classification. They introduced a model called MLP-Mixer solely based on a succession of two multilayer perceptrons applied independently to image patches and channels, respectively, which achieved comparable accuracy to the ViT~\citep{dosovitskiy2020image} on ImageNet.

Likewise, \citet{liu2021pay} argued that self-attention is not critical for computer vision and language modelling. They introduced a network called gMLP that models the interactions with Spatial Gating Units (SGU) instead of self-attention. Their model achieved the same accuracy as the ViT~\citep{dosovitskiy2020image} on ImageNet, and the same perplexity of BERT~\citep{devlin2019bert} on a subset of C4.

Alternatively, \citet{bello2021lambdanetworks} proposed to replace the Transformer's self-attention with Lambda layers. Long-range content and position-based interactions are captured by transforming the context into linear functions, i.e. matrices, and applying them to each input independently. LambdaNetworks achieved comparable results to relatively small Transformers on ImageNet classification. While the memory complexity of Lambda layers remains quadratic with respect to the sequence length, it does not scale with the batch size. Additionally, the author proposed a multi-query variant that scales down the complexity by a factor.

Finally, \citet{metaformer} argued that the architecture of the Transformer is more valuable to the performance than the specific mechanism to relate the tokens. To illustrate their claim, the authors introduced the PoolFormer, a network that performs similarly to vision Transformers while replacing the self-attention mechanism with pooling, a simple non-parametric operator. Furthermore, the authors expanded on this idea with a more general and flexible architecture called MetaFormer, where the mechanism to relate the tokens is not specified while the other components are kept the same as the Transformer.

\section{Summary of the Specialized Approaches}

\renewcommand{\arraystretch}{1.5}
\begin{table}[htb]
  \setlength\tabcolsep{15pt}
  \centering
  \caption{Summary of the specialized methods and their associated models.}
  \label{tab:specialized_methods}
  \resizebox{\textwidth}{!}{%
  \begin{tabular}{lll}
    \toprule
    Category                              & Approach                                                   & Model                                                              \\  \midrule
    \multirow{12}{*}{Sparse}              & \multirow{7}{*}{Fixed and Random Patterns}                 & Star-Transformer \citep{guo2019startransformer}                    \\
                                          &                                                            & Sparse Transformer \citep{DBLP:journals/corr/abs-1904-10509}       \\
                                          &                                                            & Cascade Transformer \citep{wang2020transformer}                    \\
                                          &                                                            & LogSparse-Transformer \citep{li2020enhancing}                      \\
                                          &                                                            & BlockBERT \citep{qiu2020blockwise}                                 \\
                                          &                                                            & Longformer \citep{2020arXiv200405150B}                             \\
                                          &                                                            & BigBird \citep{NEURIPS2020_c8512d14}                               \\ \cline{2-3}
                                          & \multirow{3}{*}{Learned and Adaptive Patterns}             & Sinkhorn Transformer \citep{tay2020sparse}                         \\
                                          &                                                            & SparseBERT \citep{shi2021sparsebert}                               \\
                                          &                                                            & Adaptively Sparse Transformer \citep{correia-etal-2019-adaptively} \\ \cline{2-3}
                                          & \multirow{2}{*}{Clustering and Locality-Sensitive Hashing} & Reformer \citep{Kitaev2020Reformer}                                \\
                                          &                                                            & Routing Transformer \citep{roy2020efficient}                       \\ \hline
    \multirow{6}{*}{Factorized Attention} & \multirow{3}{*}{Low-Rank Factorization}                    & Linformer \citep{2020arXiv200604768W}                              \\
                                          &                                                            & Synthesizers \citep{tay2020synthesizer}                            \\
                                          &                                                            & Nyströmformer \citep{xiong2021nystromformer}                       \\ \cline{2-3}
                                          & \multirow{2}{*}{Kernel Attention}                          & Linear Transformer \citep{katharopoulos2020transformers}           \\
                                          &                                                            & Performer \citep{choromanski2020rethinking}                        \\ \cline{2-3}
                                          & Clustering and Locality-Sensitive Hashing                  & Transformer with clustered attention \citep{vyas2020fast}          \\ \hline
    \multirow{3}{*}{Architectural Change} & \multirow{2}{*}{Memory}                                    & Transformer-X \citet{dai-etal-2019-transformer}                    \\
                                          &                                                            & Compressive Transformer \citet{DBLP:conf/iclr/RaePJHL20}           \\ \cline{2-3}
                                          & Sequence Compression                                       & Funnel-Transformer \citet{dai2020funneltransformer}                \\ \bottomrule
  \end{tabular}%
  }
\end{table}

\end{document}